\documentclass[10pt,twocolumn,letterpaper]{article}

\usepackage{3dv}
\usepackage{times}
\usepackage{epsfig}
\usepackage{graphicx}
\usepackage{amsmath}
\usepackage{amssymb}
\usepackage{siunitx}
\usepackage{booktabs}
\usepackage{multirow}
\usepackage[normalem]{ulem}
\usepackage{ifsym}

\DeclareMathOperator*{\argmax}{arg\,max}

\usepackage{color}
\usepackage[table]{xcolor}

\definecolor{ben}{rgb}{0.9,0.,0.5}

\definecolor{vin}{rgb}{1.0,0.5,0.0}

\definecolor{todo}{rgb}{1.0, 0., 0.}

\newcommand\blfootnote[1]{%
    \begingroup
    \renewcommand\thefootnote{}\footnote{#1}%
    \addtocounter{footnote}{-1}%
    \endgroup
}
\AtBeginDocument{
    \hypersetup{colorlinks,urlcolor=black}
}

\usepackage[pagebackref=true,breaklinks=true,colorlinks,bookmarks=false]{hyperref}

\threedvfinalcopy 

\def\threedvPaperID{357} 

\ifthreedvfinal\fi

\begin{document}
\title{R4Dyn: Exploring Radar for Self-Supervised\\Monocular Depth Estimation of Dynamic Scenes}
\author{
Stefano Gasperini$^{*,\circ,1,2}$
\qquad Patrick Koch$^{*,1,2}$
\qquad Vinzenz Dallabetta$^{2}$\\
Nassir Navab$^{1,3}$
\qquad Benjamin Busam$^{1}$
\qquad Federico Tombari$^{1,4}$\\\\
$^1$ Technical University of Munich ~\quad $^2$ BMW Group ~\quad
$^3$ Johns Hopkins University ~\quad $^4$ Google
}
\maketitle
\thispagestyle{empty}

\blfootnote{This is a preprint of the paper accepted at the International Conference on 3D Vision (3DV) 2021. \textcopyright \thinspace 2021 IEEE.}
\blfootnote{$^{*}$ The authors contributed equally.}
\blfootnote{$^{\circ}$ Contact author: Stefano Gasperini (\textit{\href{mailto:stefano.gasperini@tum.de}{stefano.gasperini@tum.de}}).}

\begin{abstract}
While self-supervised monocular depth estimation in driving scenarios has achieved comparable performance to supervised approaches, violations of the static world assumption can still lead to erroneous depth predictions of traffic participants, posing a potential safety issue. In this paper, we present R4Dyn, a novel set of techniques to use cost-efficient radar data on top of a self-supervised depth estimation framework. In particular, we show how radar can be used during training as weak supervision signal, as well as an extra input to enhance the estimation robustness at inference time. Since automotive radars are readily available, this allows to collect training data from a variety of existing vehicles. Moreover, by filtering and expanding the signal to make it compatible with learning-based approaches, we address radar inherent issues, such as noise and sparsity. With R4Dyn we are able to overcome a major limitation of self-supervised depth estimation, i.e. the prediction of traffic participants. We substantially improve the estimation on dynamic objects, such as cars by 37\% on the challenging nuScenes dataset, hence demonstrating that radar is a valuable additional sensor for monocular depth estimation in autonomous vehicles.
\end{abstract}

\section{Introduction}
\label{sec:introduction}




Depth estimation is a fundamental task for scene understanding in autonomous driving and robotics navigation.
While learning-based supervised approaches for monocular depth have achieved strong performance in outdoor scenarios~\cite{2018-fu, 2020-farooq-bhat}, the expensive LiDAR sensors required for supervision are not readily available. Additionally, collecting such ground truth data is challenging, and requires further processing, as the raw LiDAR signal may not be sufficient.

Alternative methods exploit geometrical constraints on stereo pairs or monocular videos to learn depth in a self-supervised fashion.
Image sequences offer the most inexpensive source of supervision for this task.
However, these approaches require to estimate the camera pose between frames at training time~\cite{2019-godard}, and suffer from inherent issues, such as scale ambiguity and the tendency to incorrectly estimate the depth of dynamic objects.

\begin{figure}[t]
\centering
  \includegraphics[width=0.47\textwidth]{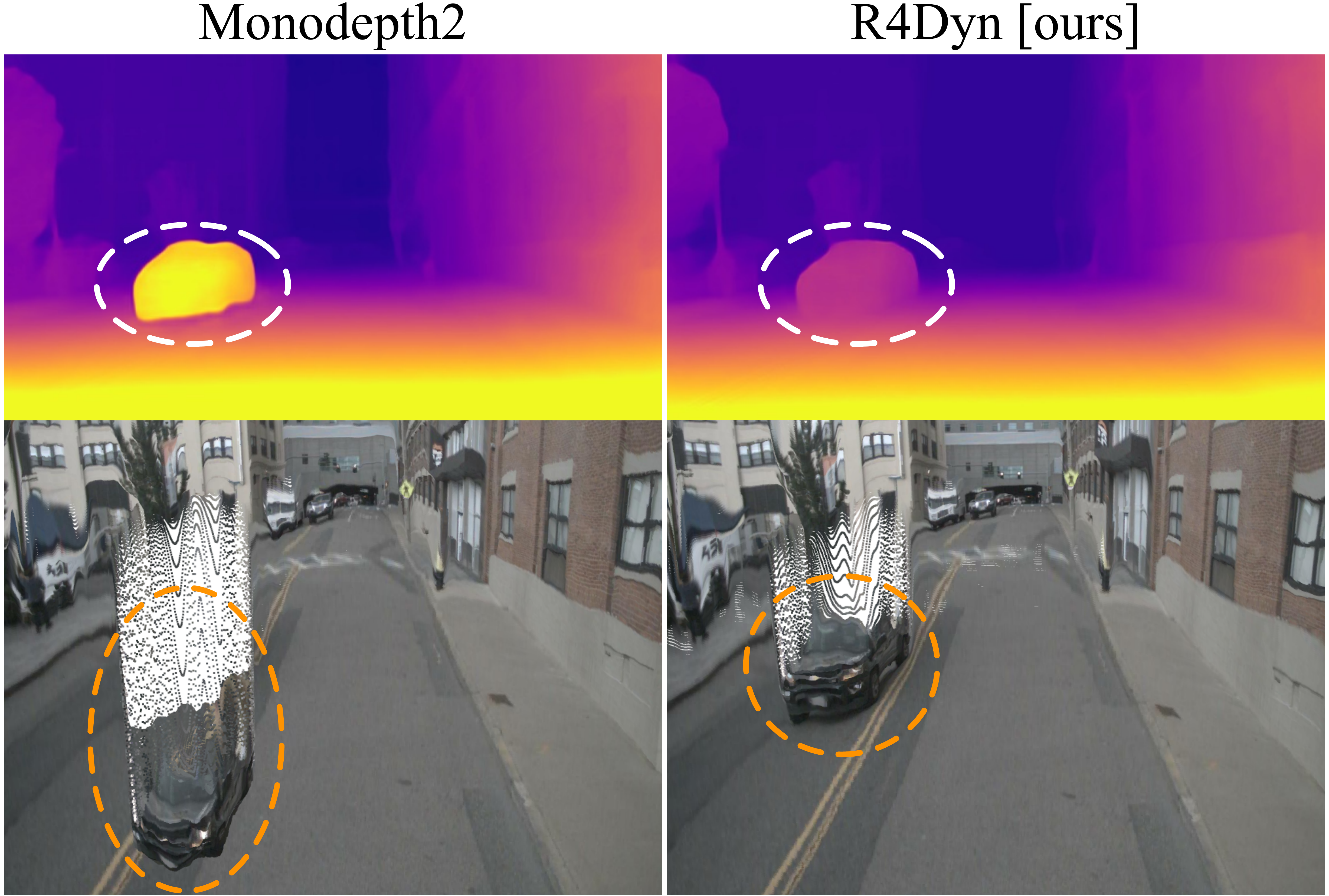}
   \caption{
   Example of depth prediction of our R4Dyn compared to that of Monodepth2~\cite{2019-godard}, from the validation set of nuScenes~\cite{2020-caesar}.
   This dynamic scene violates the static world assumption, posing a challenge for self-supervised approaches.
   The depth of the safety critical oncoming traffic is severely underestimated by Monodepth2, but correctly predicted by our R4Dyn.
   }
   \label{fig:teaser}
\end{figure}

Furthermore, self-supervised methods rely on the 
assumption of a moving camera in a static world~\cite{1979-ullman}.
As real-world street scenes are typically dynamic, this assumption is often violated, leading to significantly wrong predictions.
This raises a variety of issues, such as the infinite depth problem, where leading vehicles driving at the same speed as the camera are predicted to be infinitely far away (such as the horizon), due to their lack of relative motion across frames.
Several works addressed this problem either by excluding from the loss computation those regions where no pixel variation is detected~\cite{2019-godard}, or discarding from the training set those frames with leading vehicles driving at a similar speed~\cite{2020-guizilini2}. Others increase the complexity by estimating the individual object's motion~\cite{2019-casser} or the scene flow~\cite{2020-li}.
However, existing methods that preserve the model complexity~\cite{2019-godard, 2020-guizilini2} fail to cope with oncoming traffic, as can be seen in Figure~\ref{fig:teaser}. In this case, instead of infinitely far, the estimated depth tends to be greatly underestimated, since from the camera perspective its relative motion across frames is larger than that of static objects.

Compared to LiDARs, radars are relatively inexpensive range sensors, already integrated in a large number of mass production vehicles~\cite{2020-lin, 2021-long}, to aid features such as adaptive cruise control.
Nevertheless, radar popularity in the learning-based autonomous driving domain is yet limited, and has been explored mainly in the context of object detection~\cite{2019-nobis, 2020-kim}.
To this date and to the best of our knowledge, only two works~\cite{2020-lin, 2021-long} investigated the use of radar to improve depth estimation, both proposing a multi-network pipeline to incorporate radar data at inference time, while using LiDAR-supervision during training, transforming the task into sparse depth completion.

In this paper, we aim to bridge this gap and integrate readily available radar sensors to improve self-supervised monocular depth estimation.
As radars are already very common, our proposal allows to collect training data from a wide range of existing vehicles, instead of a few prototypes. Despite their inherent noise and sparsity, radars could provide enough information to mitigate the limitations of self-supervised approaches, overcoming the need for LiDARs.
Towards this end, we propose a novel loss formulation to complement self-supervised approaches, showing the benefits of radar as additional weak supervision signal to improve on dynamic objects.
Moreover, we optionally integrate radar data at inference time, transforming the task into very sparse depth completion.
We name our method R4Dyn (Radar for Dynamic scenes), and the contributions of this paper can be summarized as follows:
\begin{itemize}
    \item  We use radar to aid the prediction of dynamic objects in self-supervised monocular depth estimation.
    \item To the best of our knowledge, this is the first monocular depth estimation work that exploits radar as supervision signal.
    \item We propose a technique to filter and expand radar detections, and make radar compatible with learning-based solutions. 
    \item We provide extensive evaluations, including errors on safety critical dynamic objects, on the challenging nuScenes dataset~\cite{2020-caesar}, training various prior methods under equivalent settings, hence creating a new benchmark, and easing the comparisons for future works.
\end{itemize}

\begin{figure*}[t!]
\centering
  \includegraphics[width=1.00\textwidth]{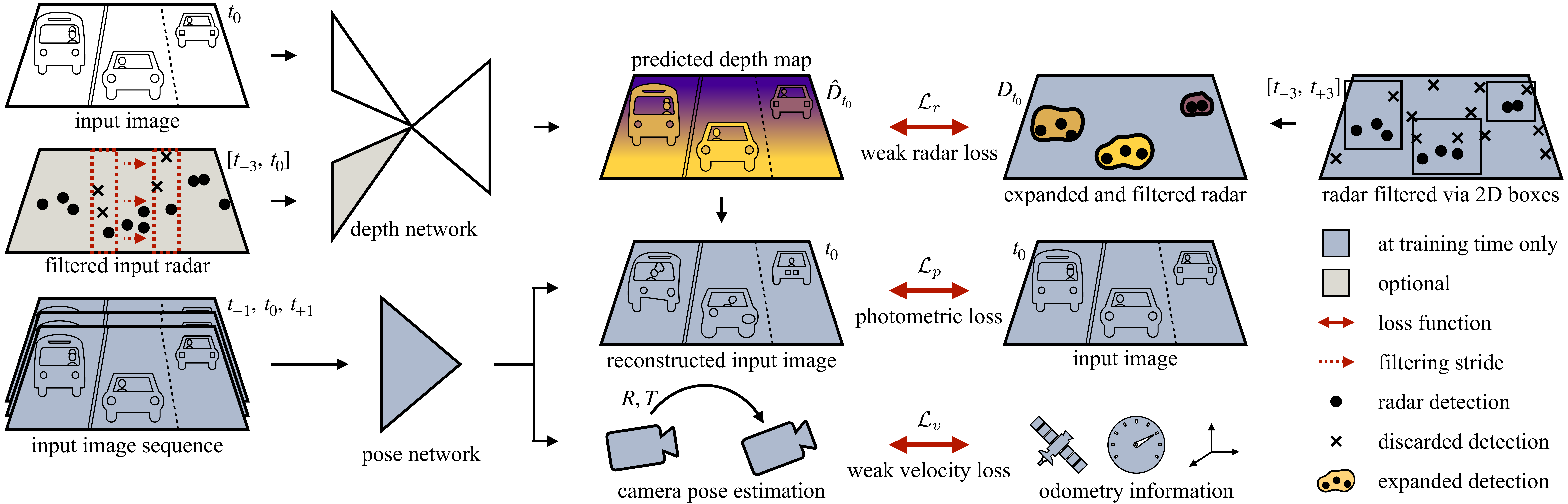}
   \caption{Overview of R4Dyn.
   The proposed approach incorporates radar as a weak supervision signal, and optionally as additional input. R4Dyn extends a self-supervised framework by incorporating radar to improve on the depth estimation of dynamic objects.
   }
   \label{fig:framework}
\end{figure*}

\section{Related Work}
\label{sec:related_work}
\subsection{Supervised Monocular Depth Estimation}

Estimating depth from a single color image is an ill-posed problem, as there is an infinite number of 3D scenes that can yield the same 2D projection. Nonetheless, tremendous advances have been achieved since Eigen et al.~\cite{2014-eigen} pioneered using CNN-based architectures and Laina et al.~\cite{2016-laina} leveraged fully-convolutional networks with residual connections~\cite{2016-kaiming} to predict dense depth maps from monocular images. While most supervised works regressed directly to the depth measurements of LiDAR sensors (as in KITTI~\cite{2013-geiger}) or RGB-D cameras (as in NYU-Depth v2~\cite{2012-silberman}), Fu et al.~\cite{2018-fu} formulated the task in an ordinal fashion.

\subsubsection{Depth Completion with Radar}
\label{subsec:related_work-radar_depth}

Despite a recently increasing interest in radar for object detection~\cite{2019-nobis, 2020-kim},
to this date, only two works~\cite{2020-lin, 2021-long} used it for depth estimation. Both achieved substantial improvements by using it in a LiDAR-supervised setting with a multi-stage architecture, with the first stage filtering the radar signal and improving its quality. In particular, by incorporating the radar as additional input, they transformed the depth estimation task into highly sparse depth completion.
Lin et al.~\cite{2020-lin} were the first to use radar in this supervised context, where they followed a late-fusion approach to account for the heterogeneity of input modalities.
Long et al.~\cite{2021-long} proposed a sophisticated learning-based association between the projected radar points and the RGB image.

Similarly to these pioneering works~\cite{2020-lin, 2021-long}, our method also incorporates radar for depth estimation, but we follow a novel idea: we focus on using radar to improve the estimation of dynamic objects in a self-supervised setting, for which we propose a specific loss function.


\subsection{Self-Supervised Monocular Depth Estimation}
\label{subsec:related_work-selfsupervised}
Self-supervised methods overcome the need for expensive LiDAR data by leveraging view reconstruction constraints, either via stereo pairs~\cite{2016-garg, 2017-godard} or monocular videos~\cite{2017-zhou, 2019-godard, 2020-guizilini-sfm}. The latter build on
the motion parallax induced by a moving camera in a static world~\cite{1979-ullman}. Furthermore, these methods require to predict simultaneously the depth and the camera pose transformation. Since the pioneering work on video-based training by Zhou et al.~\cite{2017-zhou}, vast improvements have been achieved thanks to novel loss terms~\cite{2019-godard}, detail-preserving network architectures~\cite{2020-guizilini-sfm} and the exploitation of cross-task dependencies~\cite{2018-jiao, 2020-guizilini2}.

\subsubsection{Solutions to Self-supervised Inherent Issues}

\hspace{\parindent} \textbf{Scale ambiguity}
Since infinitely many 3D objects correspond to the same 2D projection, video-based methods can only predict depth up to an unknown scale factor.
Therefore, a plethora of works~\cite{2017-zhou, 2019-godard, 2019-casser} rely on ground truth (i.e. LiDAR) median-scaling at test time. Guizilini et al.~\cite{2020-guizilini-sfm} targeted this issue by imposing a weak velocity supervision on the estimated pose transformation, exploiting the available odometry information, achieving scale-awareness.

\textbf{Dynamic scenes} Another major limitation of video-based approaches is due to the inherent static world assumption. This is perpetually violated in driving scenarios, leading to critically incorrect depth predictions of dynamic objects (e.g. traffic participants). A typical failure case is caused by leading vehicles driving at a similar speed as the ego vehicle, thereby lacking relative motion across frames and resulting in a significantly overestimated depth. Godard et al.~\cite{2019-godard} addressed this problem with an auto-masking loss to ignore pixels without relative motion.
In contrast, Guizilini et al.~\cite{2020-guizilini2} proposed a workaround to detect and discard training samples where this "infinite depth problem" occurs, thereby keeping only uncomplicated frames, albeit reducing the training data.

However, neither of the two~\cite{2019-godard, 2020-guizilini2} accounted for
oncoming traffic, which leads to a significantly underestimated depth, since its motion across frames is larger than that of the static elements.
This might be linked to the popular KITTI depth benchmark~\cite{2013-geiger} mostly lacking such kind of safety critical dynamic scenes, widely available on nuScenes~\cite{2020-caesar}.
Alternative solutions target dynamic scenes, by increasing the model complexity and simultaneously predicting depth, ego-pose, plus 3D motion of dynamic objects~\cite{2019-casser} or scene flow~\cite{2018-yang, 2020-luo, 2020-hur, 2020-li}.

Although methods learning scene flow in a self-supervised fashion do not require expensive additional labels, e.g. instance segmentation~\cite{2019-casser}, they might suffer from the same ambiguities of self-supervised depth estimation, and require stereo vision~\cite{2018-yang, 2020-luo, 2020-hur}.

Our proposed approach is substantially different from previous works targeting dynamic scenes. We address this problem by incorporating radar data, which has not been explored before in this context.








\section{Method}\label{sec:method}



In this paper we incorporate radar data to enhance the self-supervised depth prediction of dynamic scenes. An overview of our method can be seen in Figure~\ref{fig:framework}.
We build on top of a self-supervised framework which learns from monocular videos, and uses the vehicle odometry (Section~\ref{subsec:method_starting_point}), as in~\cite{2020-guizilini-sfm}, for scale-awareness.
Although radar signals are highly sparse and noisy, they could provide enough information to improve self-supervised methods.
So, we integrate radar data both during training to improve the prediction of dynamic objects (Section~\ref{subsec:method_radar_loss}), and inference to increase the overall robustness (Section~\ref{subsec:method_radar_input}).

\textbf{Radar inherent issues}
Radar and LiDAR provide significantly different signals: radar signals are noisy, sparse and often lack elevation information. Radar detections projected to the image plane occupy a single pixel, leading to a low density of $\approx 0.03\%$, compared to $\approx 1.63\%$ of a 32-beam LiDAR~\cite{2020-caesar}. Additionally, radar suffers from several sources of noise (clutter), complicating its usage in learning-based approaches. The major causes are multi-path~\cite{2021-long} and "see-through" effects, due to the different viewpoints~\cite{2020-rodriguez}, scene geometry and physical sensor properties, and not relatively simple Gaussian noise. Multi-path effects alone affect about 35\% of all points~\cite{2021-long}.
Moreover, due to missing elevation information (e.g. in nuScenes~\cite{2020-caesar}), all detections lie in a plane parallel to the ground.
This complicates radar usability, requiring new techniques to make it compatible with learning-based approaches and correctly associate radar detections with image pixels. Therefore, we tackle these issues by expanding the radar influence (Section~\ref{subsec:method_radar_loss_expansion}), and mitigating its noise (Section~\ref{subsec:method_radar_loss_filtering}).



\subsection{Self-Supervised Framework}
\label{subsec:method_starting_point}
The proposed method is built on top of a video-based monocular depth approach, described in this Section. We aim to simultaneously predict the depth $\hat{D}_t$ of a target frame and the pose transformations $T_{t \to s}$ between target $I_t$ and source frames $I_{s \in \{t-1, t+1\}}$. Depth and pose estimates serve to warp the source frames into a reconstructed target view,
from which an appearance-based error is computed from the view reconstruction~\cite{2017-zhou} and a Structural Similarity (SSIM)~\cite{2004-wang} term, as in~\cite{2019-godard, 2020-guizilini-sfm}.

Following~\cite{2019-godard},
only the minimum reprojection error $\mathcal{L}_{p}$ is considered, accounting for partial occlusions.
Moreover, pixels without relative motion across frames are masked out~\cite{2019-godard}.
An additional loss $\mathcal{L}_{s}$ encourages smoothness whilst preserving edges~\cite{2017-godard}.
$\mathcal{L}_{p}$ and $\mathcal{L}_{s}$ are computed at each scale of the depth decoder, after upsampling
to full resolution~\cite{2019-godard}.

As the radar provides absolute depth values, to use its signal as weak supervision (Section~\ref{subsec:method_radar_loss}), it is crucial to estimate depth at the right scale.
However, as the learning objectives above only allow to predict depth up to an unknown scale factor. We follow~\cite{2020-guizilini-sfm} to achieve scale-awareness with a weak velocity supervision $\mathcal{L}_{v}$ on the pose transformation.


\subsection{Weak Radar Supervision}
\label{subsec:weak_radar_supervision_SUBCHAPTER_not_the_loss}
In this work, we focus on dynamic objects, which embody safety critical failure cases of self-supervised monocular depth estimation. 
We aim to mitigate this issue, which is visible in Figure~\ref{fig:teaser}, via a weak radar-based supervision.

\begin{figure}[h]
\centering
  \includegraphics[width=0.41768021\textwidth]{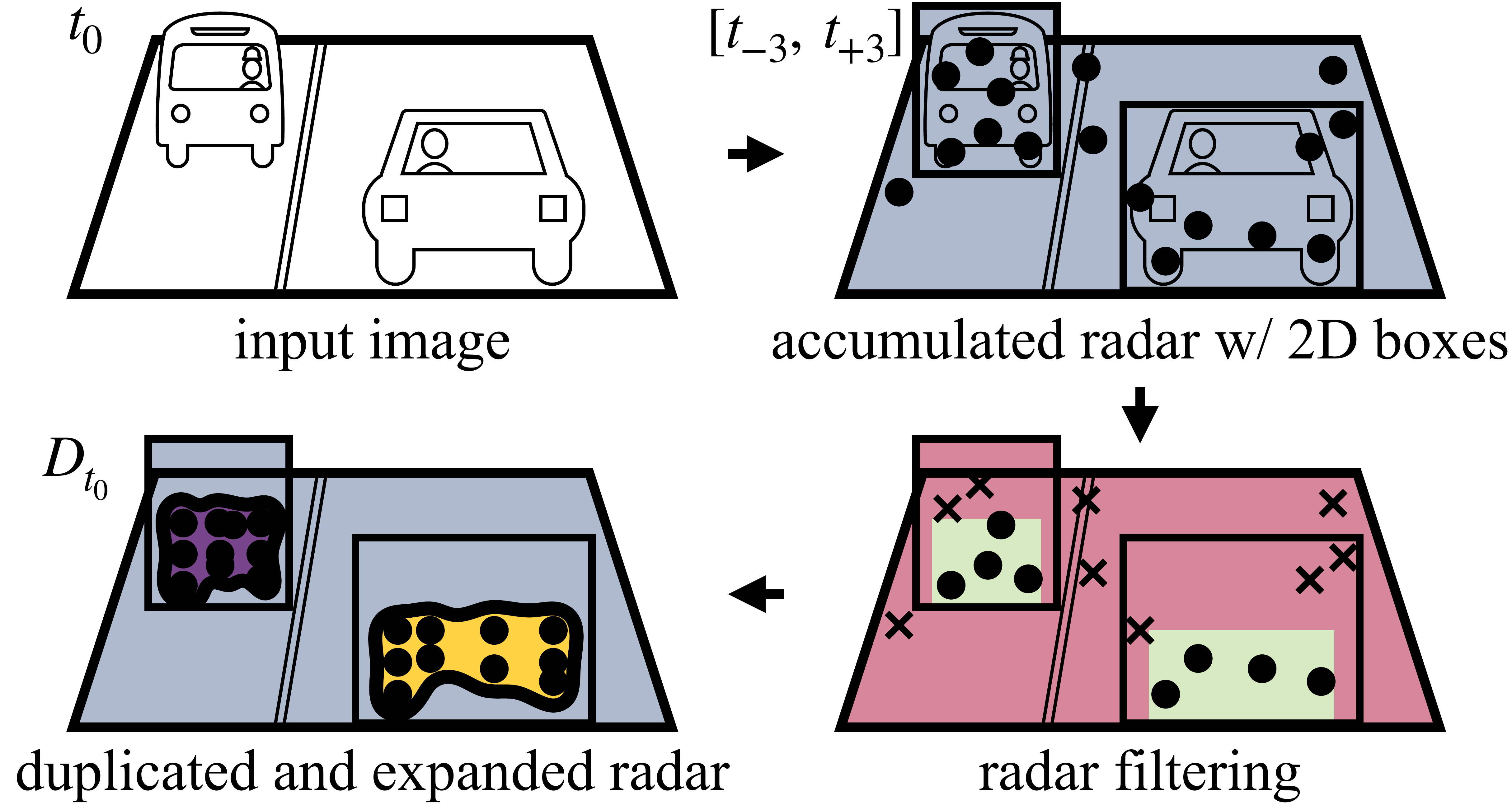}
   \caption{Radar preparation for weak supervision, detail of the top right portion of Figure~\ref{fig:framework}.
   The signal is accumulated, then filtered via 2D boxes, duplicated and expanded.
   }
   \label{fig:radar_loss_preparation}
\end{figure}
\vspace{-1.8em}

\subsubsection{Addressing Radar Sparsity}
\label{subsec:method_radar_loss_expansion}
To address radar high sparsity and make it suitable as supervision signal, it is fundamental to expand its influence over a larger portion of the image. 
In particular, as we focus on dynamic objects, we aim to expand the radar signal across all object pixels, within their boundaries.
At training time, we overcome sparsity by incorporating the RGB image context information, similarly to~\cite{1998-tomasi, 2016-barron}, as well as 2D bounding boxes.

\textbf{Context-aware radar expansion}
We expand each radar point to cover a larger area of the corresponding object.
Expanding it as much as possible, but constraining it within the boundaries, and accounting for 3D shape variations require a precise mapping between the projected radar points and the object pixels.
Towards this end, during training, we generate an association map around each radar point. We exploit the idea of bilateral filtering~\cite{1998-tomasi}, a common edge-preserving image smoothing technique.
In particular, we link an image pixel $p$ to a radar point $r$ if $p$ and $r$ are spatially close in the image space, within the same bounding box, and have similar pixel intensities (i.e. $I(p) \approx I(r)$).
This bilateral association map
can be formalized as follows:
\begin{equation}
    w(p,r) =
    \text{exp}\left(-\frac{\left(\Delta u\right)^{2}+ \left(\Delta v\right)^{2}}{2 \sigma_d^{2}}
        - \frac{\left\| \Delta I \right\|^2_2}{2 \sigma_r^{2}}\right)
\end{equation}
where $\Delta u = (u - u_r)$ and $\Delta v = (v - v_r)$ are the pixel distances, $\Delta I = I(u,v) - I(u_r, v_r)$ the difference of intensities of a radar point pixel $r = (u_r, v_r)$ and a neighboring pixel $p = (u, v)$, while $\sigma_d$ and $\sigma_r$ denote domain and range smoothing parameters, respectively. Hence, the bilateral association $w(p,r)$ represents the probability of a pixel $p$ to be linked to a radar point $r$.


\textbf{Pixel-radar association map}
Although this bilateral confidence map preserves edges by design, it may still leak small, nonzero values beyond object boundaries, leading to undesirably smooth depth predictions.~We avoid this by clipping the heatmaps in proximity of the box edges.
Moreover, $w(p,r)$ could serve as per-pixel weight for the radar supervision. However, this would put more emphasis on the radar pixel $r$ and result in uneven object depth maps.
We address this by transforming the confidence values $w(p, r)$ in a binary map. Considering the set of reliable detections $R_{df}$ (Section \ref{subsec:method_radar_loss_filtering}) and a threshold $\gamma \in \mathbb{R}$, we compute:
\begin{equation}
    W(p, r) =  
    \begin{cases}
    1 & \text{if } \max\limits_{r \in R_{df}} w(p,r) > \gamma \\
    0 & \text{otherwise}
    \end{cases}
\label{eq:binary_association_map}
\end{equation}
so all relevant portions get the same amount of supervision.

\begin{figure}[h]
\centering
  \includegraphics[width=0.47\textwidth]{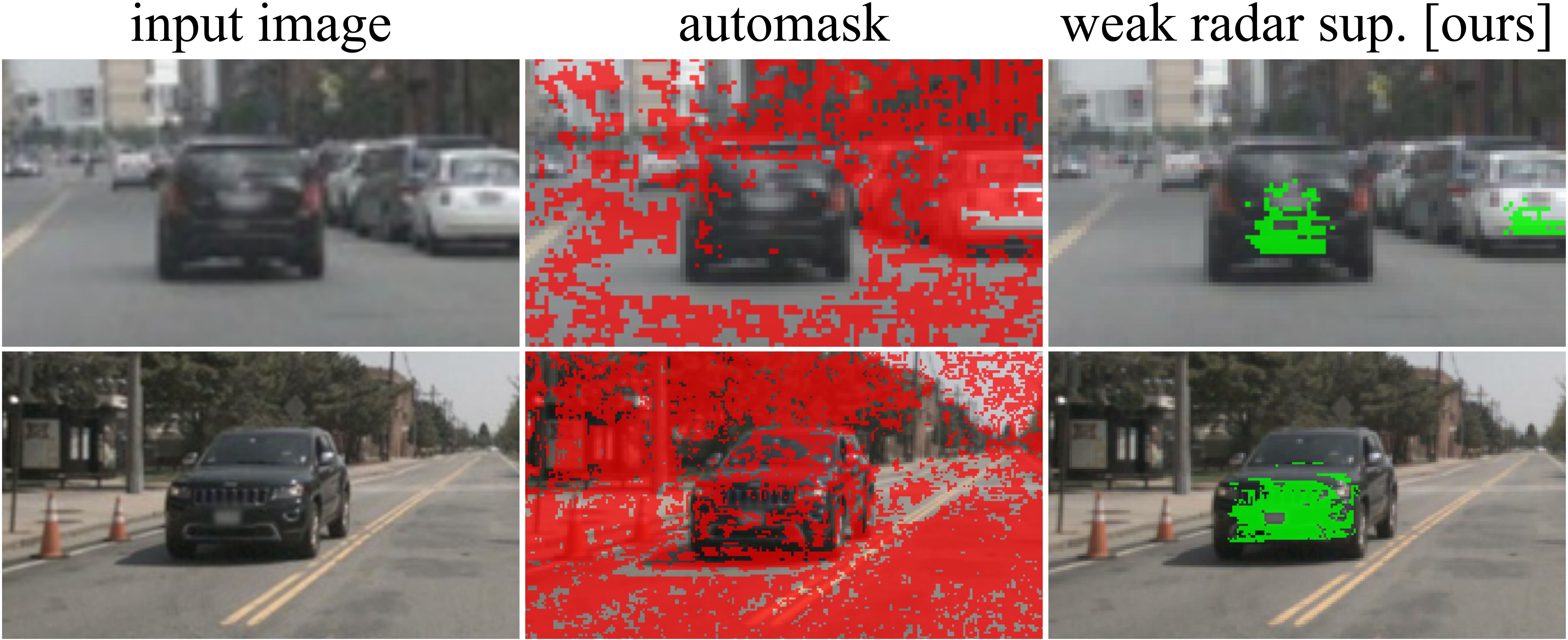}
   \caption{The automask~\cite{2019-godard} and our radar supervision mask. Red pixels contribute to the loss $\mathcal{L}_{p}$. The automask correctly masks out the leading car (\textit{top}), but not the oncoming one (\textit{bottom}). Our radar supervision $\mathcal{L}_{r}$ acts (green) successfully on both cars.
   }
   \label{fig:automask_radarsup}
\end{figure}

\textbf{Duplication}
To account for the missing elevation,
as in Figure~\ref{fig:radar_loss_preparation} we copy each projected radar point $r$ and its corresponding depth along the vertical axis to the lower third and middle of its bounding box. We leave out the upper half of the box to consider depth variations within the boxes (e.g. windshield of the car in the lower part of Figure~\ref{fig:automask_radarsup}).


\textbf{Measurement Accumulation}
\label{subsubsec:measurement_accumulation}
We also reduce sparsity by accumulating measurements.
To do so, we exploit radar doppler information to compensate for ego- and target-motion across samples, as proposed by~\cite{2020-kim}.
Since the doppler velocity is radial, it provides only a rough estimate of the true velocity of the target, but it is a reasonable approximation considering the frame rate.
Via duplication and accumulation, we obtain a dense set of radar points $R_d$.

\subsubsection{Addressing Radar Noise}
\label{subsubsec:noise_filtering}
\label{subsec:method_radar_loss_filtering}
Noise is another major radar issue to be reduced (Section~\ref{sec:method}).

\textbf{Clutter removal}
We want to extract from $R_d$ reliable measurements $R_{df}$ for supervision.
We do so by leveraging 2D bounding boxes during training to filter out noisy radar detections.
Within each object in the image space, radar detections closer to the sensor are likely to be reliable, whereas points at higher distances often result from noise. Hence, to obtain $R_{df}$, we find the minimum depth $d_m$ per bounding box $b_i$ and only keep detections within $b_i$ having $d < d_m + \beta$. The tolerance $\beta$ allows to keep multiple points in $b_i$, and accounts for depth variations along 3D objects.

\textbf{Object-focused filtering}
Radars without elevation information (e.g. nuScenes~\cite{2019-casser}) provide only detections parallel to the ground plane, thus, in the image, further points appear higher than closer ones, and within the same objects are more likely to be noisy.
Analogously, detections around the box edges could be unreliable due to wide boxes or "see-through" effects.
For these reasons, as shown in Figure~\ref{fig:radar_loss_preparation}, we discard points in the upper 50\% and outer 20\% of bounding boxes, as well as overlapping areas.
Furthermore, it is more intricate to assess radar detections reliability outside object boxes,
hence we discard all background radar points to avoid erroneous supervision.


\subsubsection{Training Objective for Dynamic Objects}
\label{subsec:method_radar_loss}
Addressed noise and sparsity, the radar signal is suitable to weakly supervise the depth prediction of dynamic objects.
We want to use it to adjust erroneous depth estimations via a loss function.
We define this weak radar supervision as:
\begin{equation}
    \mathcal{L}_{r}\left(\hat{D}, D\right) = \frac{1}{N} \sum_{p \in I} \left\| \hat{D}_p - D_p \right\|_1 \odot W(p,r)
    \label{eq:radar_loss}
\end{equation}
where $\odot$ denotes the Hadamard product. Eq.~\ref{eq:radar_loss} aims at fixing the predicted depth $\hat{D}$ towards the expanded target radar measurements $D$, for each of the $N$ image pixels $p$ where the binary association map $W(p, r)$ (Eq.~\ref{eq:binary_association_map}) is positive.

\begin{figure}[h]
\centering
  \includegraphics[width=0.47\textwidth]{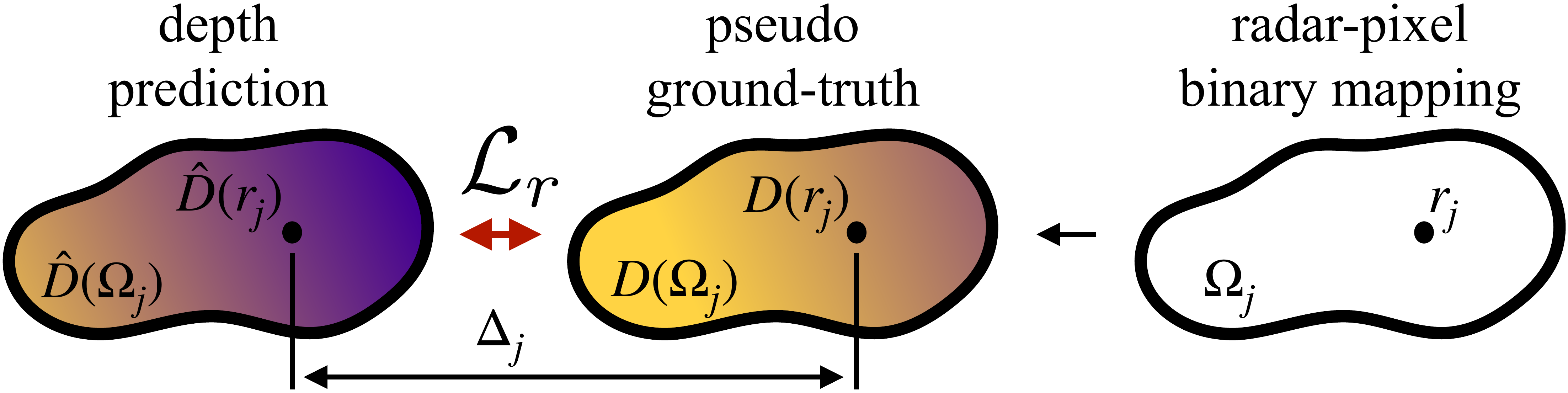}
   \caption{Depth gradient preserving mechanism.
   We compute $\Delta_j$ between prediction and radar detection, then adjust each pixel in the prediction by $\Delta_j$, by enforcing a pseudo-ground truth during training, thereby preserving the predicted gradient.
   }
   \label{fig:depth_gradient}
\end{figure}

\textbf{Depth gradient preservation}
The expanded measurements $D$ in Eq.~\ref{eq:radar_loss} act as ground truth depth.
We introduce $\Omega_j=[W(p,r_j)=1]$ as the area affected by the expanded radar point $r_j \in R_{df}$.
If $D$ was constant within each $\Omega_j$, then $\mathcal{L}_{r}$ would shift the whole area to the same depth $D$.
This would not take into account depth variations within objects, such as the door panels or the windshield of the lower car in Figure~\ref{fig:automask_radarsup} being further than its front bumper.
Moreover, although the depth of dynamic objects is often under- or over-estimated by video-based self-supervised methods, depth variations within objects are typically well predicted (e.g. in Figure~\ref{fig:teaser}).
For these reasons, we adapt the pseudo-ground truth $D$ to preserve depth variations.
As shown in Figure~\ref{fig:depth_gradient}, we do so by computing $\Delta_j=D_{radar}(r_j) - \hat{D}(r_j)$ as the difference between the prediction $\hat{D}(r_j)$ and the measurement $D_{radar}(r_j)$, only at each radar pixel $r_j \in R_{df}$.
We then generate the pseudo ground truth $D(\Omega_j)$ by shifting the prediction $\hat{D}(\Omega_j)$ by the same $\Delta_j$:
\begin{equation}
    \begin{gathered}
        D_j(p) = 
        \hat{D}_j(p) + \Delta_j
    \end{gathered}
\end{equation}
where $\Delta _j$ is computed at the radar pixel $r_j$ such that $\argmax_{r \in R_{df}} ~w(p,r_j)$.

The final objective function can thus be formulated as:
\begin{equation}
    \mathcal{L}_{} = \mathcal{L}_p + \lambda_1 \mathcal{L}_{s} + \lambda_2 \mathcal{L}_{v} + \lambda_3 \mathcal{L}_{r}
    \label{eq:final_loss}
\end{equation}
with $\lambda_1$, $\lambda_2$ and $\lambda_3$ being balancing coefficients.

\begin{table*}
\begin{center}
\begin{tabular}{l|ll|cccc|cccc}
\toprule
Method 									& Sup. 	& Input 	& AbsRel 			& SqRel 		 & RMSE 					& $\delta_1$ 	  & AbsRel$_\text{C}$ & AbsRel$_\text{V}$ 	& AbsRel$_\text{N}$ & AbsRel$_\text{P}$\\
\midrule\midrule  
Lin et al.$^\ddag$~\cite{2020-lin} 		& GT 	& ImR 	& \textbf{0.1086} 	& \textbf{1.080} & \textbf{5.394} 			& \textbf{88.21}  & 0.1907 			  & 0.2082					& 	0.2088		&	0.2930	\\
\midrule   
Struct2Depth$^{K}$~\cite{2019-casser} 		& Mi$^*$ & Im 	& 0.2195 			& 3.799	 		 & 8.441 					& 73.23 		  & 0.3323 			  & 0.3516					& 	0.3739 & 0.2993				\\
PackNet-SfM~\cite{2020-guizilini-sfm} 	& Mv 	& Im 	& 0.1567 			& 2.440 		 & 7.230 					& 82.64 		  & 0.1814 			  & 0.2382					& 	0.2508 & 0.2473				\\
Monodepth2$^\ddag$~\cite{2019-godard} 		& M$^*$ & Im 	& 0.1398 			& 1.911 		 & 6.825  					& 84.82 		  & 0.1983 			  & 0.2110					& 0.2300		& 0.2572			\\
baseline$^\ddag$~[ours] 					& Mv 	& Im 	& 0.1315 			& 1.705 		 & 6.520 					& 85.71 		  & 0.1862 			  & 0.2091					& 	0.2254		& 0.2351		\\
R4Dyn-L$^\ddag$~[ours] 					& Mvr & Im 	& 0.1296 			& \textbf{1.658} 		 & 6.536 					& 85.76 		  & 0.1343 			  & 0.1618					& 0.1686 & 0.2231					\\
R4Dyn-LI$^\ddag$~[ours] 					& Mvr & ImR 	& \textbf{0.1259} 	& 1.661 & \textbf{6.434} 			& \textbf{86.97}  & \textbf{0.1250}   & \textbf{0.1504}				 	& 	\textbf{0.1589} & \textbf{0.2146}				\\

\bottomrule
\end{tabular}
\end{center}
\caption{Evaluation on the nuScenes~\cite{2020-caesar} validation \textit{day-clear} set. $\ddag$ were pretrained on ImageNet~\cite{2009-deng}, $K$ also on KITTI~\cite{2013-geiger}. Supervisions (Sup.): GT: via LiDAR data, M: via monocular sequences, $*$: test-time median-scaling via LiDAR, i: instance masks, v: weak velocity, r: weak radar. Inputs (In.): Im: RGB, ImR: RGB and radar. C, V, N and P for \textit{Cars}, \textit{Vehicles}, \textit{Non-parked vehicles} and \textit{Pedestrians}, respectively. R4Dyn-L and R4Dyn-LI: proposed method with radar as L: weak supervision, I: input. Notation reused in other Tables.
}
\label{table:eval_all}
\end{table*}


\subsection{Sparse Depth Completion with Radar as Input} \label{subsec:radar_as_input}
\label{subsec:method_radar_input}
As radar sensors are readily available in a wide range of mass-production vehicles~\cite{2020-lin, 2021-long}, a depth estimator could exploit them at inference time, transforming the estimation task in a very sparse depth completion problem.
To do so, we first need to mitigate radar inherent sparsity and noise issues. We follow a similar approach as for using the signal as weak supervision (Section~\ref{subsec:weak_radar_supervision_SUBCHAPTER_not_the_loss}), although we do not make use of 2D bounding boxes at inference time, since extracting them for the input would increase the runtime.

For sparsity, we apply the same measurement accumulation technique described in Section~\ref{subsubsec:measurement_accumulation} to obtain a denser point cloud, aggregating the current and past radar frames.
For the noise, we adopt the same min-pooling approach as in Section~\ref{subsubsec:noise_filtering}, but instead of exploiting 2D boxes, we slide a fixed-size window across all projected radar points.

To account for the fact that radar and camera are heterogeneous sensors, we apply separate encodings and merge the features in a late fusion fashion, as in~\cite{2020-lin}.


\section{Experiments and Results}
\label{sec:experiments}

\subsection{Experimental Setup}
\label{subsec:exp_setup}
\textbf{Dataset}
We conduct all experiments on the challenging nuScenes dataset~\cite{2020-caesar}. We selected it as it is the only available large-scale public dataset with the recording vehicle fitted with a camera and an automotive radar.
NuScenes contains around 15h of driving data collected in Boston and Singapore, including diverse traffic scenarios (e.g. both left and right hand drive), with a plethora of dynamic scenes (unlike KITTI~\cite{2013-geiger}), making it difficult for self-supervised depth estimation.
As we are interested in sensor setups readily available in production cars, we consider only data from front-facing camera and radar.
For training, we use only scenes with good visibility (i.e. \textit{day-clear}).
This includes 15129 samples (with synced image, radar and LiDAR data) for the official training set and 6019 for the official validation set (of which 4449 are \textit{day-clear}).

\textbf{Evaluation metrics}
We evaluated our models on the standard depth estimation metrics and errors. As ground truth we used single raw LiDAR scans up to a maximum depth of \SI{80}{\meter}. In particular, as this work focuses on dynamic objects, we are interested in the performance improvements on such objects. Towards this end, we also evaluated according to the semantic class: we exploited LiDAR semantic segmentation annotations from~\cite{2020-caesar} to distinguish between classes, and computed the errors on the depth predictions at the corresponding LiDAR points. The evaluated classes comprise \textit{Cars}, \textit{Vehicles} (e.g. \textit{Cars}, \textit{Trucks}, \textit{Buses}, \textit{Motorcycles}, \textit{Bicycles}), \textit{Non-parked Vehicles} and \textit{Pedestrians}, thereby encompassing all traffic participants. 

\textbf{Network architecture}
The presented techniques for processing and incorporating radar data for depth are not tailored to a specific backbone architecture.
We used the small and effective ResNet-18-based~\cite{2016-kaiming} pose and depth networks from~\cite{2019-godard}. The pose network comprised 13.0M parameters, while the depth network 14.8M. With the radar as input, an additional ResNet-18 encoder branch was used, increasing the depth parameters by 1.1M.

\begin{table}[h]
\begin{center}
\begin{tabular}{ll|cc}
\toprule
ID & Weak radar sup. & AbsRel & AbsRel$_\text{C}$\\
\midrule\midrule
L1 & baseline & 0.1315 & 0.1862 \\  
L2 & L1 + $\mathcal{L}_{r}$ {w/} raw pts  & 0.1766 & 0.4841 \\ 
L3 & L2 + filter pts {w/} GT box  & 0.1323 & 0.1960 \\ 
L4 & L3 + bilateral expansion & 0.1306 & 0.1551 \\ 
L5 & L4 + binary mapping & 0.1297 & 0.1356 \\ 
L6 & L5 + depth gradient pres.  & 0.1296 & \textbf{0.1343} \\
L7 & L6 -- GT + predicted box & \textbf{0.1289} & \textbf{0.1343} \\ 
\bottomrule
\end{tabular}
\end{center}
\caption{Weak radar supervision ablation, evaluated on the nuScenes~\cite{2020-caesar} \textit{day-clear} validation set. The baseline L1 is defined in Section~\ref{subsec:method_starting_point}.
L3-L6 and L7 make use of 2D boxes from ground truth annotations or an off-the-shelf detector~\cite{2021-wang}, respectively. L6 is equivalent to R4Dyn-L, with image-only input (Im).}
\label{table:ablation_loss_function}
\end{table}



\textbf{Implementation details}
Each training sample consisted of a 576x320 resolution image triplet ($t_{-1}, t_0, t_{+1}$), 4 radar measurements at $[t_{-3}, t_{0}]$ for the input, and 7 radar samples for the supervision at training time at $[t_{-3}, t_{+3}]$. At inference time we used a single image at $t_{0}$, and 4 radar measurements at $[t_{-3}, t_{0}]$.
Using the Adam optimizer~\cite{2015-kingma} with $\beta_1= 0.9$, $\beta_2= 0.999$, we trained with a batch size of 16 for a total of 40 epochs with initial learning rate \num{2e-4}, which was halved every 10 epochs. If enabled, we introduced the radar supervision after 30 epochs, when the correct depth scale had already been learned, and set the learning rate to \num{1e-5}, halved after 8 epochs. The loss balancing weights were set to $\lambda_1 = \num{1e-3}$, $\lambda_2 = \num{0.02}$, $\lambda_3 = \num{0.2}$. During training, we applied random horizontal flipping to the input data as well color jittering with brightness, contrast, saturation $\pm 0.2$ and hue $\pm 0.05$ to the input images.
The window for filtering radar at inference time was 320 pixels tall and 8 wide, with stride 3, while the tolerance $\beta$ was $2m$.
Further details can be found in the supplementary material.
In the following, we refer to different configurations of our method, namely R4Dyn-L, R4Dyn-I, and R4Dyn-LI, with -L and -I denoting our weak radar supervision and radar as input, respectively. 
We trained all models using PyTorch on a single NVIDIA Tesla V100 32GB GPU. Inference of our full method (i.e. R4Dyn-LI) takes 27ms on an NVIDIA GTX 1080 8GB GPU.


\begin{table}[h]
\begin{center}
\begin{tabular}{cl|cc}
\toprule
ID & Input config. & AbsRel & AbsRel$_\text{C}$ \\
\midrule\midrule
I1 & RGB only     & 0.1315 & 0.1862 \\  
I2 & I1 + 1 radar input    & 0.1357 & 0.1926 \\ 
I3 & I2 + $\mathcal{L}_{r}$   & 0.1298 & 0.1319 \\ 
I4 & I3 + radar accum.  & 0.1301 & 0.1279 \\ 
I5 & I4 + doppler accum. & 0.1273 & 0.1264 \\ 
I6 & I5 + filtered & \textbf{0.1259} & \textbf{0.1250} \\ 
\bottomrule
\end{tabular}
\end{center}
\caption{Pre-processing ablation for radar as input, evaluated on the \textit{day-clear} nuScenes~\cite{2020-caesar} validation set. I1 is defined in Section~\ref{subsec:method_starting_point}. I6 is equivalent to R4Dyn-LI, our full approach.}
\label{table:input_ablation}
\end{table}

\textbf{Prior works and baseline}
For a fair comparison, we retrained all methods on the same dataset split, using the same image resolution, the official implementations and parameters, until convergence. For Struct2Depth~\cite{2019-casser} we started from model weights pretrained on ImageNet~\cite{2009-deng} and then KITTI~\cite{2013-geiger} by the authors, which improved its convergence over ImageNet-pretraining only.
All methods except for PackNet-SfM~\cite{2020-guizilini-sfm} used a ResNet-18~\cite{2016-kaiming} backbone pretrained on ImageNet~\cite{2009-deng}.
Our baseline is the self-supervised image-only method described in Section~\ref{subsec:method_starting_point}, which includes the weak velocity supervision.


\begin{figure*}[t]
\centering
  \includegraphics[width=1.00\textwidth]{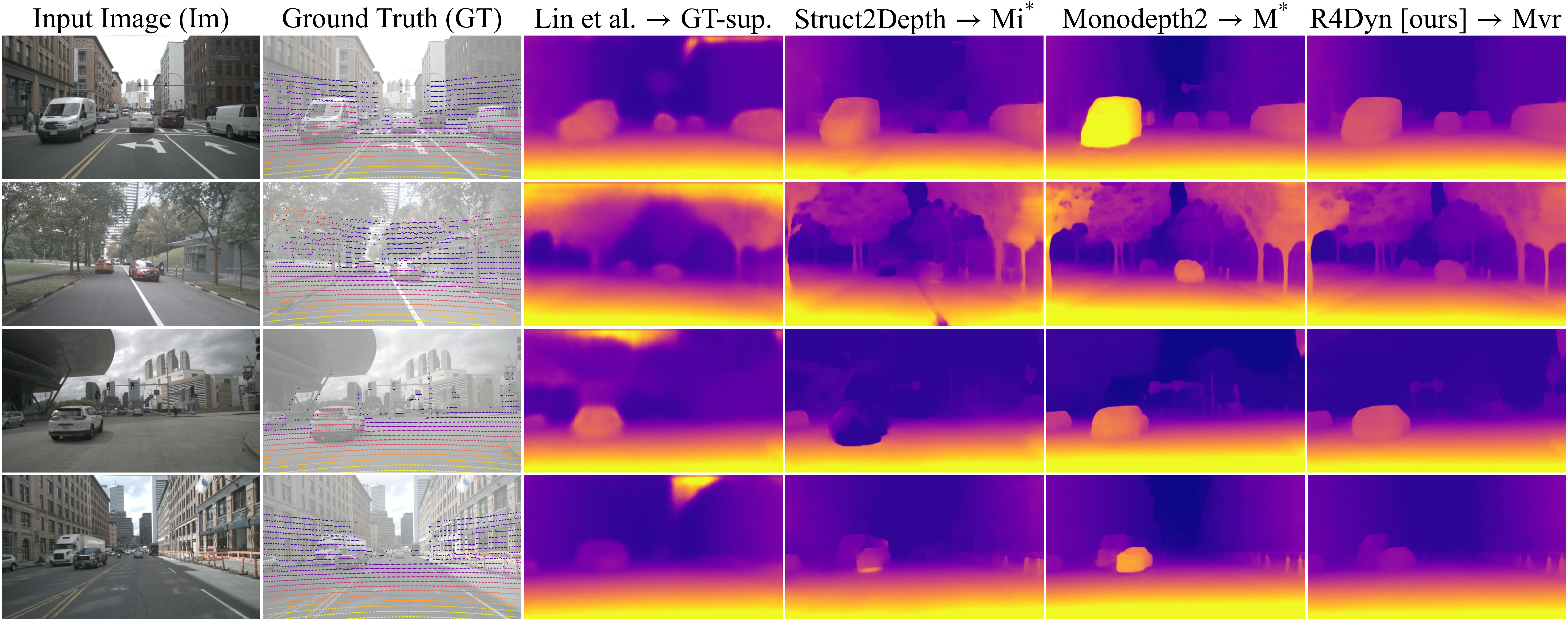}
   \caption{Qualitative results from related works on the nuScenes~\cite{2020-caesar} validation set. Next to each method name we indicate its supervision.}
   \label{fig:method_comparison}
\end{figure*}

\begin{table}[h]
\begin{center}
\begin{tabular}{r|cccc}
\toprule
Radar \% & AbsRel & AbsRel$_\text{C}$  & AbsRel$_\text{V}$ & AbsRel$_\text{P}$ \\
\midrule\midrule
0~\%  & 0.1315 & 0.1862 & 0.2091 & 0.2351 \\  
25~\%  & 0.1309  & 0.1358 & 0.1672 & 0.2254 \\ 
50~\%  & 0.1261  & 0.1309 & 0.1553 & 0.2167 \\ 
100~\%  & \textbf{0.1259} & \textbf{0.1250} & \textbf{0.1504} & \textbf{0.2146} \\ 
\bottomrule
\end{tabular}
\end{center}
\caption{Evaluation on the nuScenes~\cite{2020-caesar} \textit{day-clear} validation set. Different amounts of radar points are shown, both for input and weak supervision. 0\% is the baseline, while 100\% is R4Dyn-LI. C, V and P stand for \textit{Cars}, \textit{Vehicles} and \textit{Pedestrians} respectively.}
\label{table:radar_percentage}
\end{table}

\subsection{Quantitative Results}\label{subsec:quantitative_results}

\textbf{Comparison with related methods}
Table~\ref{table:eval_all} shows the comparison between our R4Dyn and related approaches. We report alternative solutions for dynamic objects, such as Struct2Depth~\cite{2019-casser} and Monodepth2~\cite{2019-godard}, as well as another method using radar for depth, i.e. the LiDAR-supervised work by Lin et al.~\cite{2020-lin}. 
Our R4Dyn-L and R4Dyn-LI outperformed the strong baseline across the board, by a significant margin. Remarkably, the error on \textit{Cars} dropped by a substantial 33\% with R4Dyn-LI, showing the benefit of radar for estimating the depth of dynamic objects. Our approach largely improved over Monodepth2~\cite{2019-godard}, on which our baseline builds upon. As can also be seen in Figure~\ref{fig:automask_radarsup}, the automask from Monodepth2~\cite{2019-godard} was not able to cope with oncoming traffic, occurring frequently in the nuScenes dataset. 
This led to a large error on \textit{Cars} and other traffic participants. Our R4Dyn-L and R4Dyn-LI reduced it by 32\% and 37\% respectively.

\begin{table}[h]
\begin{center}
\begin{tabular}{l|cccc}
\toprule
Method & \textit{all} & \textit{clear} & \textit{rain} & \textit{night} \\
\midrule\midrule
Lin et al.~\cite{2020-lin} & \textbf{0.126} & \textbf{0.109} & \textbf{0.145} & 0.248 \\
Struct2Depth~\cite{2019-casser} & 0.238 & 0.220 & 0.271 & 0.336 \\
PackNet-SfM~\cite{2020-guizilini-sfm} & 0.168 & 0.157 & 0.177 & 0.262 \\
Monodepth2~\cite{2019-godard} & 0.161 & 0.140 & 0.193 & 0.287\\
baseline~[ours] & 0.146 & 0.132 & 0.166 & 0.242\\
R4Dyn-L~[ours] & 0.147 & 0.130 & 0.164 & 0.273 \\
R4Dyn-LI~[ours] & 0.137 & 0.126 & 0.146 & \textbf{0.219}\\
\bottomrule
\end{tabular}
\end{center}
\caption{Evaluation of AbsRel to generalize on adverse and unseen conditions of the validation set of nuScenes~\cite{2020-caesar}. All methods were trained on \textit{day-clear} (\textit{clear}).}
\label{table:generalizability}
\end{table}


Struct2Depth~\cite{2019-casser}, despite being additionally pretrained on KITTI~\cite{2013-geiger} and its individual motion predictions, produced the worst estimations, with large inconsistencies (AbsRel std. 0.1511). Our R4Dyn-LI outputs were far more consistent (std. 0.079). We attribute this difference to the superiority of radar over instance masks as weak supervision, as well as to Struct2Depth~\cite{2019-casser} not fully solving the infinite depth problem (e.g. in Figure~\ref{fig:method_comparison}). 
The sophisticated PackNet architecture~\cite{2020-guizilini-sfm} was not able to deliver satisfactory results, which could be due to the larger model size (129M instead of 15M for ResNet-18), and the relatively small dataset. This motivated using ResNet~\cite{2016-kaiming} as backbone.
Furthermore, the LiDAR-supervised work by Lin et al.~\cite{2020-lin}, with radar as input, performed better than ours overall, albeit worse on safety critical traffic participants, such as by 53\% on \textit{Cars}. This could be due to the sparsity of the LiDAR from which it learned. 
Overall, Table~\ref{table:eval_all} demonstrates the benefit of radar for monocular depth estimation, as it can substantially improve the predictions of safety critical dynamic objects, both as weak supervision with R4Dyn-L and as input with R4Dyn-LI.

\textbf{Weak Radar supervision ablation study}
Table~\ref{table:ablation_loss_function} shows the impact of the various components of our weak radar supervision. Throughout the table, errors do not decrease with the introduction of each and every feature, thereby confirming that the radar signal necessitates to be filtered (row L3) and expanded (L4) before it can positively contribute over the baseline (L1), while simply integrating the raw points (L2) increased the errors.
With L7, we show that 2D bounding boxes can be extracted via an off-the-shelf detector, such as Scaled-YOLOv4~\cite{2021-wang} trained on MS COCO~\cite{2014-lin}, removing the need for ground truth annotations. L7 improved over L6 (which used ground truth boxes), presumably due to the nuScenes~\cite{2020-caesar} boxes being oversized (used in L3 to L6).
Overall, Table~\ref{table:ablation_loss_function} confirms the importance of our modifications to use radar as weak supervision.

\textbf{Radar as input ablation study}
Table~\ref{table:input_ablation} reports various configurations for using radar as input, again motivating the expansion (I4 and I5) and filtering (I6) of the radar signal. In fact, simply adding a single radar sweep in input (I2) did not outperform the baseline (I1).

\textbf{Variable amount of radar signal}
In Table~\ref{table:radar_percentage} we show how the output quality changed by excluding a variable amount of radar detections from the dataset. Already 25\% of the radar detections brought a large improvement over the baseline (0\%), overall as well as on safety critical traffic participants. Once again, this shows the benefit of radar for depth estimation, despite its inherent noise and sparsity. In particular, adding more detections (filtered and expanded), systematically reduced all the errors.
Moreover, considering the rapid progressing of sensor technology~\cite{2019-mart-sensor-review}, higher resolution (e.g. 200\%, 400\%) and less noisy automotive radar sensors might be available in the future, which would further increase the gap to the RGB-only baseline.

\textbf{Generalization to unseen adverse conditions}
Table~\ref{table:generalizability} reports AbsRel of our R4Dyn and related methods on difficult unseen conditions. This shows the ability of each to generalize to different data distributions. Our R4Dyn-LI outperformed all self- and weakly-supervised approaches in all conditions. Compared to the LiDAR-supervised work by Lin et al.~\cite{2020-lin}, R4Dyn-LI performs similarly in rain and significantly better in night scenes, reiterating the effectiveness and robustness of our techniques.


\subsection{Qualitative Results}
Qualitative results in Figure~\ref{fig:method_comparison} confirm the findings of our experiments, showing the superiority of our R4Dyn in estimating the depth of traffic participants.
In particular, Struct2Depth~\cite{2019-casser} had frequent issues with leading vehicles (first 3 scenes), and added a halo effect to the oncoming car in the fourth scene.
Monodepth2~\cite{2019-godard} was able to correctly estimate leading vehicles, thanks to its automask, but not oncoming traffic (as seen in Figure~\ref{fig:teaser} and inspected in Figure~\ref{fig:automask_radarsup}), which resulted in severe underestimations (first, second and fourth scene).
The LiDAR-supervised work by Lin et al.~\cite{2020-lin} correctly estimated the overall depth, but missed most details, delivering blurred outputs.
Instead, our R4Dyn could accurately estimate all challenging dynamic scenes, preserving sharp details.

\section{Conclusion}

In this paper we proposed R4Dyn, a set of techniques to integrate radar into a self-supervised monocular depth estimation framework. Extensive experiments showed the benefit of using radar both during training as weak supervision, and at inference time as added input. Our method substantially improved on the prediction of safety critical traffic participants over all related works. Therefore, R4Dyn constitutes a valuable step towards robust depth estimation. Additionally, the inexpensive and readily available setup required, allows to collect training data from a variety of existing vehicles, removing the need for expensive LiDAR data.

\appendix
\section{Supplementary Material}
In this Section we include further details and additional results computed from the same models reported in the main manuscript.

\subsection{LiDAR and Radar}
\begin{figure}[h]
\centering
  \includegraphics[width=0.47\textwidth]{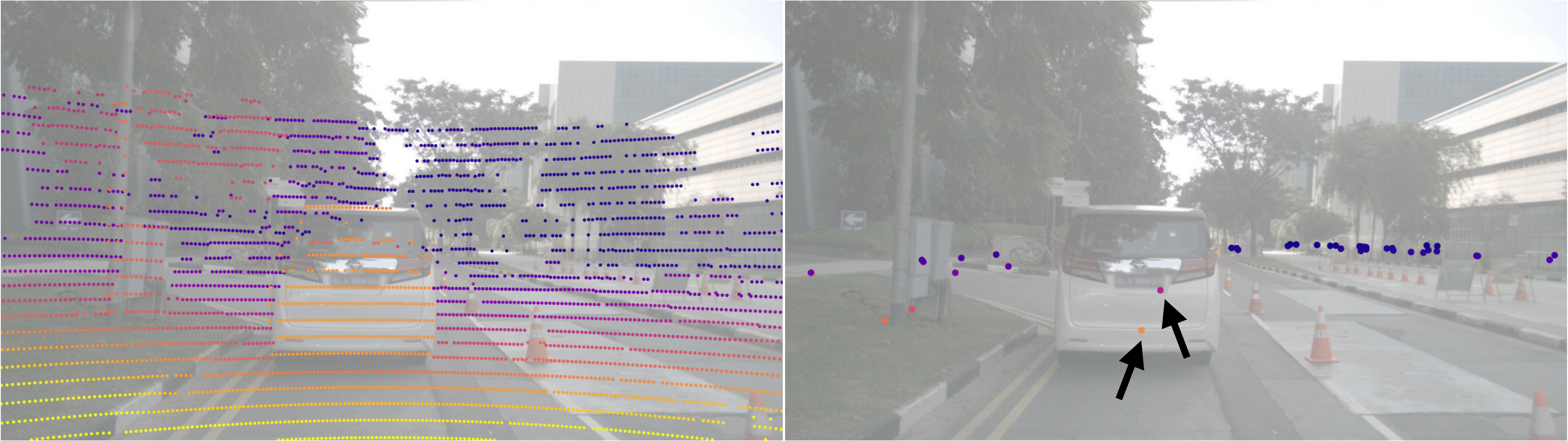}
   \caption{LiDAR (\textit{left}) and radar (\textit{right}) signals from nuScenes~\cite{2020-caesar}. The two radar measurements marked by arrows are 5.2 meters apart, due to multi-path effects.}
   \label{fig:radar_lidar}
\end{figure}

In Figure~\ref{fig:radar_lidar} we show the significant difference between LiDAR and radar signals. As described in Section~\ref{sec:method}, due to the physical properties of the sensor, radar signals are noisy, sparse and often 1-dimensional.
The complex multi-path noise can be seen affecting the two points marked by arrows in the Figure. Despite their mutual distance of 5.2 meters, the points are projected onto the same foreground object in the image space. This once again shows the importance of our modifications to handle radar data for weak supervision (Section~\ref{subsec:method_radar_loss}).

\subsection{Self-supervised Framework}
In the following, we further specify the loss functions described in Section~\ref{subsec:method_starting_point}. The photometric loss is a combination of $\mathcal{L}_1$-loss and SSIM~\cite{2004-wang}, as in~\cite{2017-godard}:
\begin{equation}
\begin{gathered}
    \mathcal{L}_1 (I_t, \hat{I}_t) = \left\| I_t - \hat{I}_t\right\|_1 \\
    \mathcal{L}_{\textup{SSIM}} = 1 - \textup{SSIM}\left(I_t, \hat{I}_t\right)\\
    pe\left(I_t, \hat{I}_t\right) = \left(1-\alpha\right) \mathcal{L}_1 (I_t, \hat{I}_t) + \frac{\alpha}{2} \mathcal{L}_{\textup{SSIM}}(I_t, \hat{I}_t)
\label{eq:photometric_error}
\end{gathered}
\end{equation}
where $\alpha = 0.85$ balances between the $\mathcal{L}_1$ and the SSIM term. Additionally, as in~\cite{2019-godard}, we only consider the minimum reprojection error to account for partial occlusions:

\begin{equation}
    \mathcal{L}_{p}\left(I_t, \hat{I}_{s \to t}\right) = \min_{s} pe\left(I_t, \hat{I}_{s \to t}\right).
\label{eq:minimum_reprojection_loss}
\end{equation}

Furthermore, we follow~\cite{2019-godard} by automatically masking out pixels which do not change appearance in between frames:
\begin{equation}
    M_{a} = \min_{s} \mathcal{L}_{p}\left(I_t, I_s\right) > \min_{s} \mathcal{L}_{p}\left(I_t, \hat{I}_{s \to t}\right).
\label{eq:automask}
\end{equation}

Hence, the photometric loss is only considered in regions where $M_a = 1$. Moreover, to encourage local smoothness while preserving edges we use a specific term from~\cite{2017-godard}:
\begin{equation}
    \mathcal{L}_{s}\left(I_t, d_t^*\right) = \frac{1}{N} \sum_{p \in N} \sum_{i \in x,y} \left | \partial_i d_t^* (p) \right | e^{-\lvert \partial_i I_t \rvert}
\label{eq:smoothness}
\end{equation}
where $\lvert \cdot \rvert$ denotes the element-wise absolute value, $\partial_x$ and $\partial_y$ are gradients in x- and y-direction, and $d_t^* = d_t^*/\overline{d_t^*}$ the mean-normalized inverse of the depth prediction.

As described in Section~\ref{subsec:method_starting_point}, we follow~\cite{2020-guizilini-sfm} with a weak velocity supervision $\mathcal{L}_{v}$ for scale-awareness:
\begin{equation}
    \mathcal{L}_{v}\left({\hat{\mathbf{t}}_{t \to s}},\mathbf{t}_{t \to s}\right)
        = \Bigl|\lVert \hat{\mathbf{t}}_{t \to s} \rVert_2 - \lVert \mathbf{t}_{t \to s} \rVert_2\Bigr|
\label{eq:velsup}
\end{equation}
where $\hat{\mathbf{t}}_{t \to s}$ and $\mathbf{t}_{t \to s}$ are the predicted and ground truth pose translations respectively, easily obtainable from readily available velocity information (e.g. via odometry).

\begin{figure}[h]
\centering
  \includegraphics[width=0.47\textwidth]{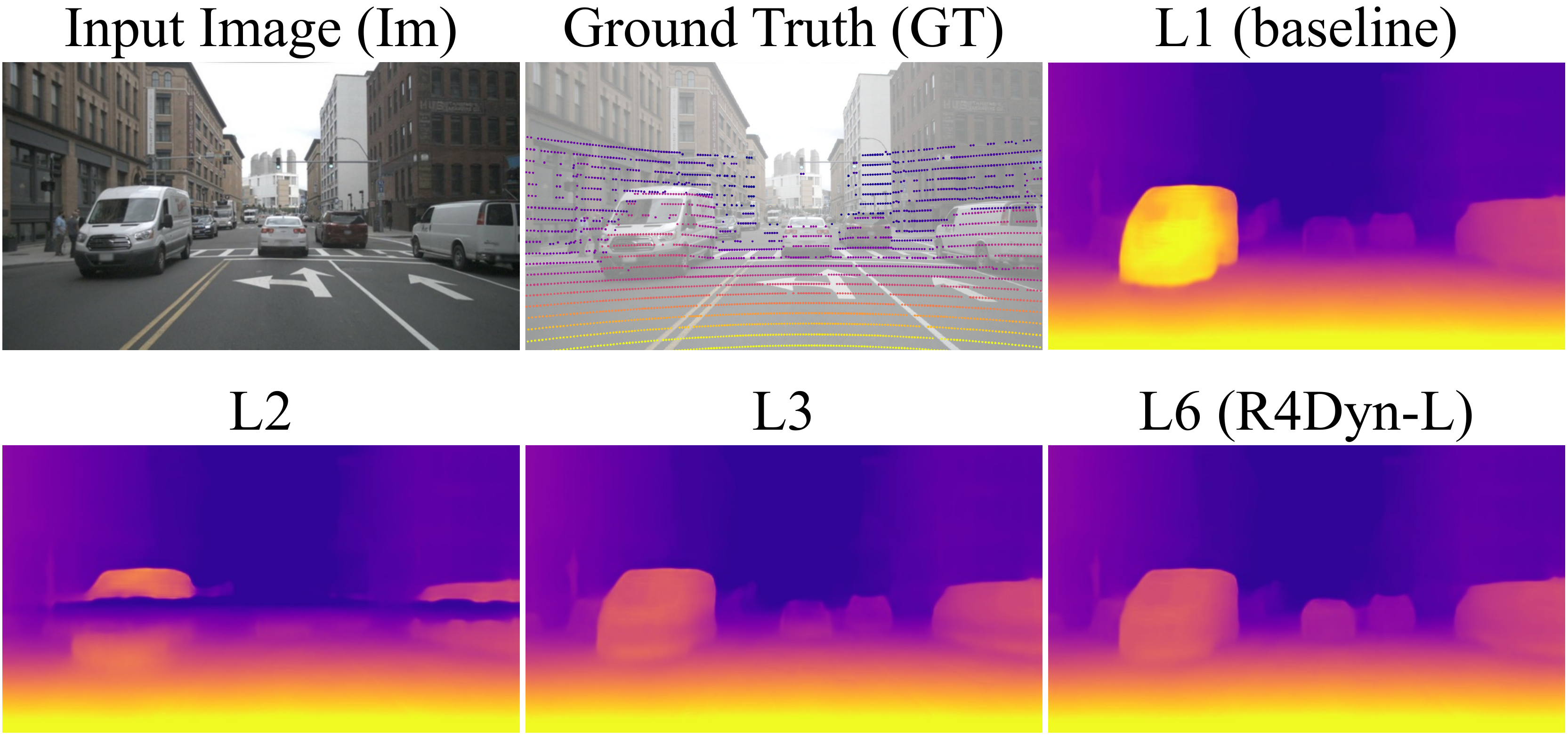}
   \caption{Qualitative comparison complementing the ablation study on the weak radar supervision (Table~\ref{table:ablation_loss_function}).}
   \label{fig:qual_loss_ablation}
\end{figure}
\vspace{-1em}
\begin{figure}[h]
\centering
  \includegraphics[width=0.47\textwidth]{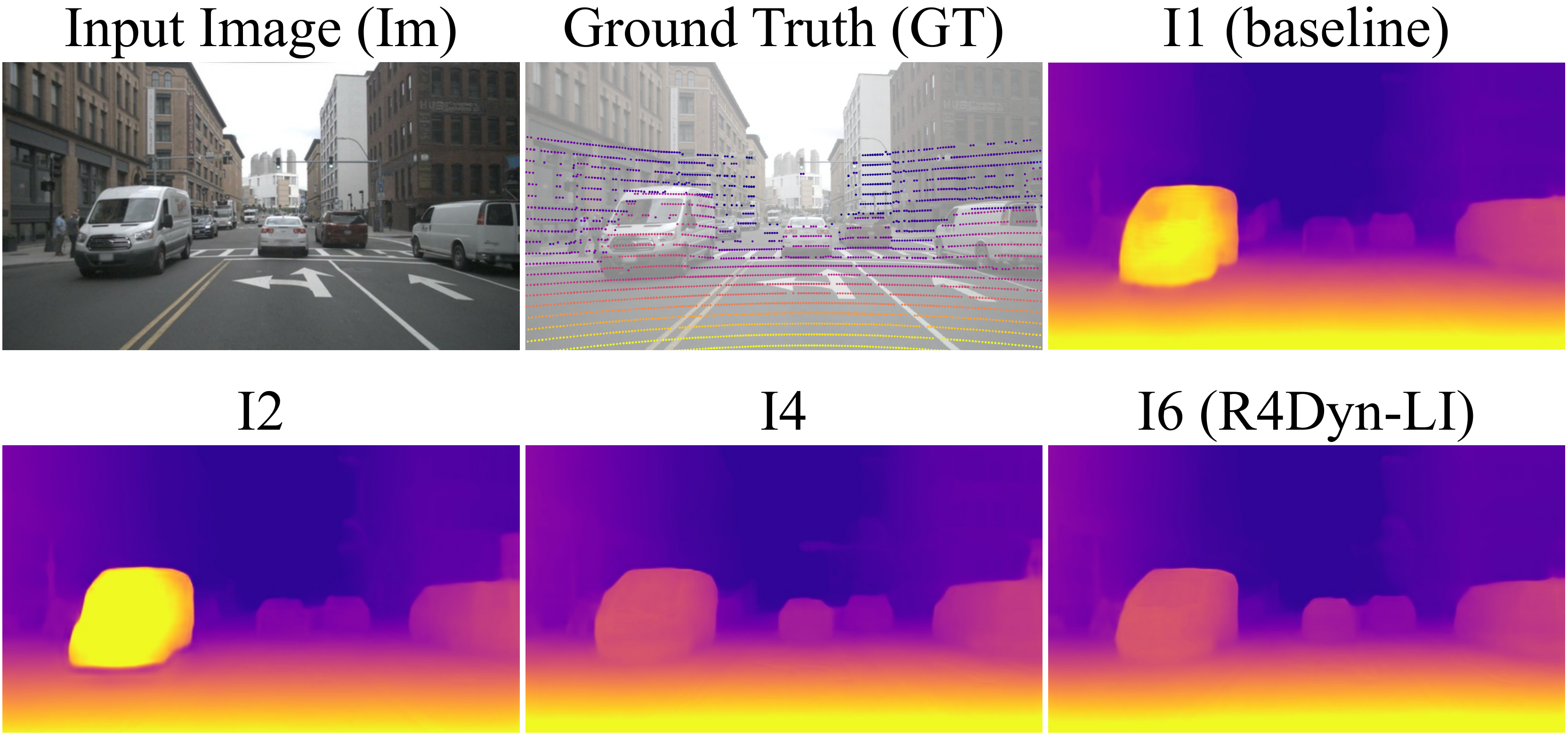}
   \caption{Qualitative comparison complementing the ablation study on the pre-processing for radar as additional input (Table~\ref{table:input_ablation}).}
   \label{fig:qual_input_ablation}
\end{figure}

\begin{table}[b]
\begin{center}
\begin{tabular}{l|cccc}
\toprule
Method & \textit{all} & \textit{clear} & \textit{rain} & \textit{night} \\
\midrule\midrule
\multicolumn{5}{c}{\textit{Cars}}\\
\midrule
Lin et al.~\cite{2020-lin} & 0.217 & 0.191 & 0.225 & 0.433 \\
Struct2Depth~\cite{2019-casser} & 0.346 & 0.332 & 0.320 & 0.535 \\
PackNet-SfM~\cite{2020-guizilini-sfm} & 0.219 & 0.181 & 0.270 & 0.449\\
Monodepth2~\cite{2019-godard} & 0.232 & 0.199 & 0.261 & 0.486 \\
baseline [ours] & 0.212 & 0.186 & 0.240 & 0.391 \\
R4Dyn-L [ours] & 0.164 & 0.134 & 0.219 & 0.311\\
R4Dyn-LI [ours] & \textbf{0.140} & \textbf{0.125} & \textbf{0.171} & \textbf{0.204}\\

\midrule
\multicolumn{5}{c}{\textit{Vehicles}}\\
\midrule
Lin et al.~\cite{2020-lin} & 0.229 & 0.208 & 0.233 & 0.428 \\
Struct2Depth~\cite{2019-casser} & 0.361 & 0.352 & 0.335 & 0.535 \\
PackNet-SfM~\cite{2020-guizilini-sfm} & 0.259 & 0.238 & 0.274 & 0.440 \\
Monodepth2~\cite{2019-godard} & 0.240 & 0.211 & 0.270 & 0.483 \\
baseline [ours] & 0.231 & 0.209 & 0.259 & 0.384 \\
R4Dyn-L [ours] & 0.182 & 0.162 & 0.220 & 0.308\\
R4Dyn-LI [ours] & \textbf{0.159} & \textbf{0.150} & \textbf{0.176} & \textbf{0.200}\\

\midrule
\multicolumn{5}{c}{\textit{Pedestrians}}\\
\midrule
Lin et al.~\cite{2020-lin} & 0.292 & 0.293 & 0.254 & 0.401 \\
Struct2Depth~\cite{2019-casser} & 0.298 & 0.299 & 0.276 & 0.367 \\
PackNet-SfM~\cite{2020-guizilini-sfm} & 0.252 & 0.247 & 0.284 & 0.355
 \\
 Monodepth2~\cite{2019-godard} & 0.258 & 0.257 & 0.253 & \textbf{0.317} \\
baseline [ours] & 0.238 & 0.235 & 0.244 & 0.348 \\
R4Dyn-L [ours] & 0.229 & 0.223 & 0.255 & 0.409 \\
R4Dyn-LI [ours] &\textbf{0.218} & \textbf{0.215} & \textbf{0.234} & 0.332 \\
\bottomrule
\end{tabular}
\end{center}
\caption{Evaluation of AbsRel on \textit{Cars}, \textit{Vehicles} and \textit{Pedestrians} under adverse conditions of the validation set of nuScenes~\cite{2020-caesar}. All methods were trained only on scenes with \textit{day-clear} (\textit{clear}) conditions. This Table complements Table~\ref{table:generalizability}, therefore shows the ability of each method to generalize to unseen settings.}
\label{table:generalizability_vehicles}
\end{table}

\begin{table*}
\begin{center}
\begin{tabular}{c|l|ll|ccccccc}
\toprule
Cl. & Method & Sup. & Input & AbsRel & SqRel & RMSE & RMSE$_{log}$ & $\delta_1$ & $\delta_2$ & $\delta_3$ \\
\midrule\midrule
\parbox[t]{2mm}{\multirow{7}{*}{\rotatebox[origin=c]{90}{\textit{Cars}}}}
&Lin et al.~\cite{2020-lin} & GT & ImR & 0.1907 & 2.399 & 6.922 & 0.2460 & 75.71 & 91.02 & 96.06\\
&Struct2Depth~\cite{2019-casser} & M$^*$ & Im & 0.3323 & 7.436 & 9.353 & 0.3307 & 57.38 & 80.54 & 89.98 \\
&PackNet-SfM~\cite{2020-guizilini-sfm} & Mv & Im &0.1814 & 1.936 & 6.313 & 0.2341 & 72.88 & 91.02 & 96.46 \\
&Monodepth2~\cite{2019-godard}&  M$^*$ & Im & 0.1983 & 2.100 & 6.635 & 0.2509 & 68.23 & 88.76 & 94.81 \\
&baseline~[ours]  & Mv & Im &  0.1862 & 2.115 & 6.735 & 0.2495 & 70.41 & 87.96 & 94.41 \\
&R4Dyn-L~[ours]  & Mvr & Im &0.1343 & 1.481 & 5.713 & 0.1913 & 81.45 & 94.02 & 97.32 \\
&R4Dyn-LI~[ours]  & Mvr & ImR  & \textbf{0.1250} & \textbf{1.371} & \textbf{5.395} & \textbf{0.1813} & \textbf{84.14} & \textbf{94.38} & \textbf{97.43} \\

\midrule
\parbox[t]{2mm}{\multirow{7}{*}{\rotatebox[origin=c]{90}{\textit{NP-Cars}}}}
&Lin et al.~\cite{2020-lin} & GT & ImR & 0.1898 & 2.485 & 6.793 & 0.2418 & 77.15 & 90.69 & 95.65 \\
&Struct2Depth~\cite{2019-casser} & M$^*$ & Im & 0.3703 & 9.063 & 10.05 & 0.3472 & 54.42 & 78.15 & 88.35 \\
&PackNet-SfM~\cite{2020-guizilini-sfm} & Mv & Im & 0.1902 & 2.050 & 6.397 & 0.2394 & 70.56 & 90.35 & 96.20 \\
&Monodepth2~\cite{2019-godard} & M$^*$ & Im & 0.2130 & 2.368 & 6.905 & 0.2624 & 65.05 & 87.03 & 93.97 \\
&baseline~[ours]  &  Mv & Im & 0.1862 & 2.115 & 6.735 & 0.2495 & 70.41 & 87.96 & 94.41 \\
&R4Dyn-L~[ours] & Mvr & Im & 0.1356 & 1.518 & 5.651 & 0.1894 & 80.92 & 93.56 & 97.16 \\
&R4Dyn-LI~[ours]  & Mvr & ImR & \textbf{0.1274} & \textbf{1.399} & \textbf{5.302} & \textbf{0.1793} & \textbf{83.80} & \textbf{94.01} & \textbf{97.43} \\

\midrule
\parbox[t]{2mm}{\multirow{7}{*}{\rotatebox[origin=c]{90}{\textit{Buses}}}}
&Lin et al.~\cite{2020-lin}& GT & ImR & 0.2330 & \textbf{3.336} & 8.456 & 0.2728 & 65.46 & 85.48 & 93.44 \\
&Struct2Depth~\cite{2019-casser} & M$^*$ & Im &0.3962 & 7.487 & 12.11 & 0.4057 & 41.22 & 69.19 & 85.83 \\
&PackNet-SfM~\cite{2020-guizilini-sfm} & Mv & Im &0.3434 & 7.328 & 11.60 & 0.3662 & 52.96 & 77.66 & 88.94 \\ 
&Monodepth2~\cite{2019-godard} & M$^*$ & Im &0.2442 & 4.518 & 9.781 & 0.3007 & 63.13 & 83.44 & 91.84 \\
&baseline~[ours]  & Mv & Im & 0.2547 & 4.614 & 9.586 & 0.3080 & 62.85 & 82.28 & 91.73 \\
&R4Dyn-L~[ours] & Mvr & Im & 0.2187 &	3.950 &	8.719 &	0.2779 &	68.08 &	85.45 &	93.14 \\
&R4Dyn-LI~[ours]  & Mvr & ImR  & \textbf{0.2055} & 3.706 & \textbf{8.316} & \textbf{0.2600} & \textbf{70.35} & \textbf{86.28} & \textbf{93.46} \\

\midrule
\parbox[t]{2mm}{\multirow{7}{*}{\rotatebox[origin=c]{90}{\textit{Trucks}}}}
&Lin et al.~\cite{2020-lin}& GT & ImR & \textbf{0.2356} & \textbf{3.248} & \textbf{8.410} & \textbf{0.2829} & \textbf{63.70} & \textbf{83.46} & \textbf{92.67} \\
&Struct2Depth~\cite{2019-casser} & M$^*$ & Im &0.3711 & 7.153 & 12.26 & 0.4027 & 45.32 & 71.41 & 84.74 \\
&PackNet-SfM~\cite{2020-guizilini-sfm} & Mv & Im &0.3472 & 6.561 & 11.48 & 0.3758 & 49.82 & 75.84 & 87.82 \\ 
&Monodepth2~\cite{2019-godard} & M$^*$ & Im &0.2659 & 5.330 & 10.42 & 0.3221 & 60.76 & 81.03 & 91.48 \\
&baseline~[ours]  & Mv & Im & 0.2751 & 4.782 & 10.25 & 0.3340 & 55.26 & 80.55 & 90.05 \\
&R4Dyn-L~[ours] & Mvr & Im & 0.2457 & 4.184 & 9.739 & 0.3077 & 58.71 & 83.26 & 92.09 \\
&R4Dyn-LI~[ours]  & Mvr & ImR  &0.2369 & 4.219 & 9.493 & 0.2997 & 63.17 & 82.85 & 92.64 \\
\midrule

\parbox[t]{2mm}{\multirow{7}{*}{\rotatebox[origin=c]{90}{\textit{Motorcycles}}}}
&Lin et al.~\cite{2020-lin}& GT & ImR & 0.2529 & 3.473 & 8.292 & 0.2757 & 62.84 & 84.27 & \textbf{93.50} \\
&Struct2Depth~\cite{2019-casser} & M$^*$ & Im &0.2328 & 3.110 & 8.026 & 0.2854 & 58.39 & 82.58 & 90.00 \\
&PackNet-SfM~\cite{2020-guizilini-sfm} & Mv & Im &0.2007 & 2.575 & 7.062 & 0.2529 & 67.38 & 88.33 & 93.28 \\
&Monodepth2~\cite{2019-godard} & M$^*$ & Im &0.1900 & 2.457 & 6.693 & 0.2409 & 72.86 & 89.33 & 93.23 \\
&baseline~[ours]  & Mv & Im & 0.1849 & 2.481 & 6.856 & 0.2486 & 72.11 & 89.26 & 92.93 \\
&R4Dyn-L~[ours] & Mvr & Im & 0.1833 & 2.534 & 6.885 & 0.2496 & 69.71 & 89.43 & 93.05 \\
&R4Dyn-LI~[ours]  & Mvr & ImR  & \textbf{0.1730} & \textbf{2.401} & \textbf{6.519} & \textbf{0.2389} & \textbf{75.26} & \textbf{89.56} & 92.64 \\
\bottomrule
\end{tabular}
\end{center}
\caption{Class(Cl.)-wise evaluation on the main individual \textit{Vehicle} classes on the nuScenes~\cite{2020-caesar} \textit{day-clear} validation set. \textit{NP} stands for \textit{Non-Parked}. Table to be considered in conjunction with Table~\ref{tab:class_eval_remaining}.}
\label{tab:class_eval_vehicles}
\end{table*}


\begin{table*}
\begin{center}
\begin{tabular}{c|l|ll|ccccccc}
\toprule
Cl. & Method & Sup. & Input & AbsRel & SqRel & RMSE & RMSE$_{log}$ & $\delta_1$ & $\delta_2$ & $\delta_3$ \\
\midrule\midrule

\parbox[t]{2mm}{\multirow{7}{*}{\rotatebox[origin=c]{90}{\textit{Vehicles}}}}
&Lin et al.~\cite{2020-lin}& GT & ImR & 0.2082 & 2.637 & 7.400 & 0.2668 & 73.08 & 88.99 & 95.08 \\
&Struct2Depth~\cite{2019-casser} & M$^*$ & Im &0.3516 & 7.179 & 10.13 & 0.3612 & 54.53 & 79.25 & 89.63 \\
&PackNet-SfM~\cite{2020-guizilini-sfm} & Mv & Im &0.2382 & 3.393 & 7.927 & 0.2885 & 67.48 & 87.71 & 94.14 \\
&Monodepth2~\cite{2019-godard} & M$^*$ & Im &0.2110 & 2.809 & 7.617 & 0.2726 & 68.89 & 87.85 & 94.50 \\
&baseline~[ours]  & Mv & Im & 0.2091 & 2.680 & 7.597 & 0.2775 & 67.91 & 87.33 & 93.91 \\
&R4Dyn-L~[ours] & Mvr & Im & 0.1618 & 2.047 & 6.681 & 0.2273 & 77.16 & 92.72 & \textbf{96.76} \\
&R4Dyn-LI~[ours]  & Mvr & ImR  &\textbf{0.1504} & \textbf{1.922} & \textbf{6.371} & \textbf{0.2188} & \textbf{80.51} & \textbf{92.77} & 96.75 \\
\midrule
\parbox[t]{2mm}{\multirow{7}{*}{\rotatebox[origin=c]{90}{\textit{NP-Vehicles}}}}
&Lin et al.~\cite{2020-lin}& GT & ImR & 0.2088 & 2.892 & 7.396 & 0.2597 & 74.98 & 89.46 & 94.96 \\
&Struct2Depth~\cite{2019-casser} & M$^*$ & Im &0.3739 & 8.440 & 10.58 & 0.3643 & 53.17 & 77.87 & 88.58 \\
&PackNet-SfM~\cite{2020-guizilini-sfm} & Mv & Im &0.2508 & 3.699 & 8.044 & 0.2923 & 66.24 & 86.47 & 93.39 \\
&Monodepth2~\cite{2019-godard} & M$^*$ & Im &0.2300 & 3.372 & 7.917 & 0.2815 & 66.25 & 86.16 & 93.29 \\
&baseline~[ours]  & Mv & Im & 0.2254 & 3.219 & 7.885 & 0.2880 & 66.67 & 84.78 & 92.27 \\
&R4Dyn-L~[ours] & Mvr & Im & 0.1686 & 2.461 & 6.728 & 0.2268 & 77.86 & 91.87 & 95.97 \\
&R4Dyn-LI~[ours]  & Mvr & ImR  & \textbf{0.1589} & \textbf{2.311} & \textbf{6.375} & \textbf{0.2162} & \textbf{80.86} & \textbf{92.61} & \textbf{96.16} \\
\midrule

\parbox[t]{2mm}{\multirow{7}{*}{\rotatebox[origin=c]{90}{\textit{Pedestrians}}}}
&Lin et al.~\cite{2020-lin} & GT & ImR & 0.2930 & 4.496 & 9.507 & 0.2966 & 59.51 & 84.11 & \textbf{93.09} \\
&Struct2Depth~\cite{2019-casser} & M$^*$ & Im & 0.2993 & 5.489 & 11.48 & 0.3714 & 49.37 & 74.98 & 87.35 \\
&PackNet-SfM~\cite{2020-guizilini-sfm} & Mv & Im & 0.2473 & 4.171 & 9.301 & 0.2987 & 61.93 & 86.17 & 92.33 \\
&Monodepth2~\cite{2019-godard} & M$^*$ & Im & 0.2572 & 4.420 & 9.831 & 0.3087 & 61.46 & 86.04 & 91.88 \\
&baseline~[ours]  & Mv & Im & 0.2351 & 4.004 & 9.117 & 0.2961 & 66.20 & 86.37 & 92.08 \\
&R4Dyn-L~[ours] & Mvr & Im &0.2231 & 3.670 & 8.806 & 0.2853 & 66.90 & \textbf{87.00} & 92.42  \\
&R4Dyn-LI~[ours] & Mvr & ImR & \textbf{0.2146} & \textbf{3.613} & \textbf{8.560} & \textbf{0.2763} & \textbf{70.74} & 86.73 & 92.41 \\

\midrule
\parbox[t]{2mm}{\multirow{7}{*}{\rotatebox[origin=c]{90}{\textit{Objects}}}} 
&Lin et al.~\cite{2020-lin} & GT & ImR & 0.2227 & 2.767 & 7.319 & 0.2735 & 72.33 & 88.68 & 95.10 \\
&Struct2Depth~\cite{2019-casser} & M$^*$ & Im & 0.3400 & 6.552 & 9.726 & 0.3543 & 56.16 & 80.38 & 90.92 \\
&PackNet-SfM~\cite{2020-guizilini-sfm} & Mv & Im & 0.2383 & 3.405 & 7.757 & 0.2861 & 69.02 & 88.50 & 94.75 \\
&Monodepth2~\cite{2019-godard} & M$^*$ & Im & 0.2123 & 2.832 & 7.592 & 0.2700 & 70.11 & 89.14 & 95.37 \\
&baseline~[ours]  & Mv  & Im & 0.2032 & 2.548 & 7.278 & 0.2693 & 70.90 & 88.97 & 95.11 \\
&R4Dyn-L~[ours] & Mvr & Im & 0.1631 & \textbf{1.997} & 6.522 & 0.2279 & 78.42 & \textbf{93.22} & \textbf{97.16}  \\
&R4Dyn-LI~[ours]  & Mvr & ImR & \textbf{0.1551} & 2.020 & \textbf{6.367} & \textbf{0.2222} & \textbf{81.20} & 92.73 & 97.09 \\

\midrule
\parbox[t]{2mm}{\multirow{7}{*}{\rotatebox[origin=c]{90}{\textit{Other}}}}
&Lin et al.~\cite{2020-lin}& GT & ImR & \textbf{0.0781} & \textbf{0.693} & \textbf{4.268} & \textbf{0.1542} & \textbf{93.07} & \textbf{97.35} & \textbf{98.73} \\
&Struct2Depth~\cite{2019-casser} & M$^*$ & Im &0.1873 & 3.254 & 6.823 & 0.2645 & 78.61 & 90.80 & 95.42 \\
&PackNet-SfM~\cite{2020-guizilini-sfm} & Mv & Im &0.1141 & 1.723 & 5.777 & 0.2051 & 88.68 & 95.54 & 97.70 \\
&Monodepth2~\cite{2019-godard} & M$^*$ & Im & 0.1117 & 1.344 & 5.354 & 0.1913 & 89.50 & 96.54 & 98.15 \\
&baseline~[ours]  & Mv & Im & 0.1010 & 1.207 & 5.223 & 0.1916 & 90.73 & 96.32 & 97.97 \\
&R4Dyn-L~[ours] & Mvr & Im & 0.1036 & 1.194 & 5.290 & 0.1931 & 90.20 & 96.21 & 97.93 \\
&R4Dyn-LI~[ours]  & Mvr & ImR  &0.1003 & 1.197 & 5.190 & 0.1886 & 91.13 & 96.41 & 98.00 \\
\bottomrule
\end{tabular}
\end{center}
\caption{
Class(Cl.)-wise evaluation on \textit{Vehicles} (includes all individual classes from Table~\ref{tab:class_eval_vehicles}, plus \textit{Bicylces}, \textit{Construction Vehicles} and \textit{Emergency Vehicles}), \textit{Non-Parked Vehicles}, \textit{Pedestrians}, \textit{Objects} (all object classes together) and \textit{Other} (pixels that do not correspond to objects) on the nuScenes~\cite{2020-caesar} \textit{day-clear} validation set. Table to be considered in conjunction with Table~\ref{tab:class_eval_vehicles}.
}
\label{tab:class_eval_remaining}
\end{table*}

\subsection{Radar Accumulation}
The accumulation of multiple radar measurements described in Section~\ref{subsubsec:measurement_accumulation} is performed by mapping points from time $t-\tau$ to $t$, as proposed by~\cite{2020-kim}. Specifically, the ego- and target-motion compensation across multiple measurements is done considering the doppler velocity as:
\begin{equation}
\begin{aligned}
    &x_t = x_{t-\tau} + \Delta d_{x,t} + v_{x,t-\tau} \cdot \Delta \tau \\
    &y_t = y_{t-\tau} + \Delta d_{y,t} + v_{y, t-\tau} \cdot \Delta \tau \\
    &z_t = z_{t-\tau} + \Delta d_{z,t}
    \label{eq:doppler_motion_estimation}
\end{aligned}
\end{equation}
where $x_t$, $y_t$ and $z_t$ denote the estimated target location at time $t$, $\Delta d$ the ego pose transformation between measurements, and $v$ the measured doppler velocity of the target.

\subsection{Additional Implementation Details}
As we focus on safety critical dynamic objects (e.g. traffic participants), when filtering the radar signal to use it as weak supervision (Section~\ref{subsec:weak_radar_supervision_SUBCHAPTER_not_the_loss}), we only consider 2D bounding boxes of the following classes of nuScenes~\cite{2020-caesar}: \textit{Vehicles} (including all sub-classes) and \textit{Pedestrians} (including all sub-classes). As nuScenes provides only 3D bounding boxes, we obtained 2D boxes by considering the top, left, bottom, and right extremes of the projected 3D boxes. Furthermore, as we want to expand the radar signal over the objects, the binary association mask from Equation~\ref{eq:binary_association_map} should vary in size depending on the object, e.g. larger for a bus, than for a pedestrian. Additionally, we consider as most reliable detections those projected in the lower center of a bounding box. Towards this end, we set the smoothing parameter $\sigma_d$ according to the bounding box dimensions, hence accounting for the object size: we set $\sigma_{d,x} = c \cdot s \cdot bb_w/2$ and $\sigma_{d,y} = c \cdot s \cdot bb_h/2$, where $bb_w$ and $bb_h$ are the box width and height, and $c$ is a constant scale factor set to $1.5$.
Moreover, $s$ is a scale factor that depends on the position of the considered radar point with respect to the bounding box: $s = \delta_{side}/(bb_w/2) \cdot \delta_{top}/bb_h$, with $\delta_{side}$ and $\delta_{top}$ being the minimum distance of the radar point from the bounding box side and top edges respectively. The range smoothing parameter is fixed to $\sigma_r=\num{1e-5}$.

For prior works, as mentioned in Section~\ref{subsec:exp_setup}, we used the original implementation and parameters, adapted as follows to accomodate the different dataset (i.e. nuScenes~\cite{2020-caesar}). We trained Monodepth2~\cite{2019-godard} for 40 epochs, which is the same as our R4Dyn. PackNet-SfM~\cite{2020-guizilini-sfm} was trained for a total of 200 epochs due its to slow convergence, and Struct2Depth~\cite{2019-casser} was trained for 75 epochs after the full KITTI~\cite{2013-geiger} training performed by the authors. For fairness when comparing with other approaches, for Struct2Depth~\cite{2019-casser} we used the motion model (denoted by the authors with an M), but not their online refinement (indicated with an R). To ensure fair comparability with other methods, for the work of Lin et al.~\cite{2020-lin} we only used radar measurements from the past and present, but not from the future, which were used in the original implementation.

\begin{figure}[h]
\centering
  \includegraphics[width=0.49\textwidth]{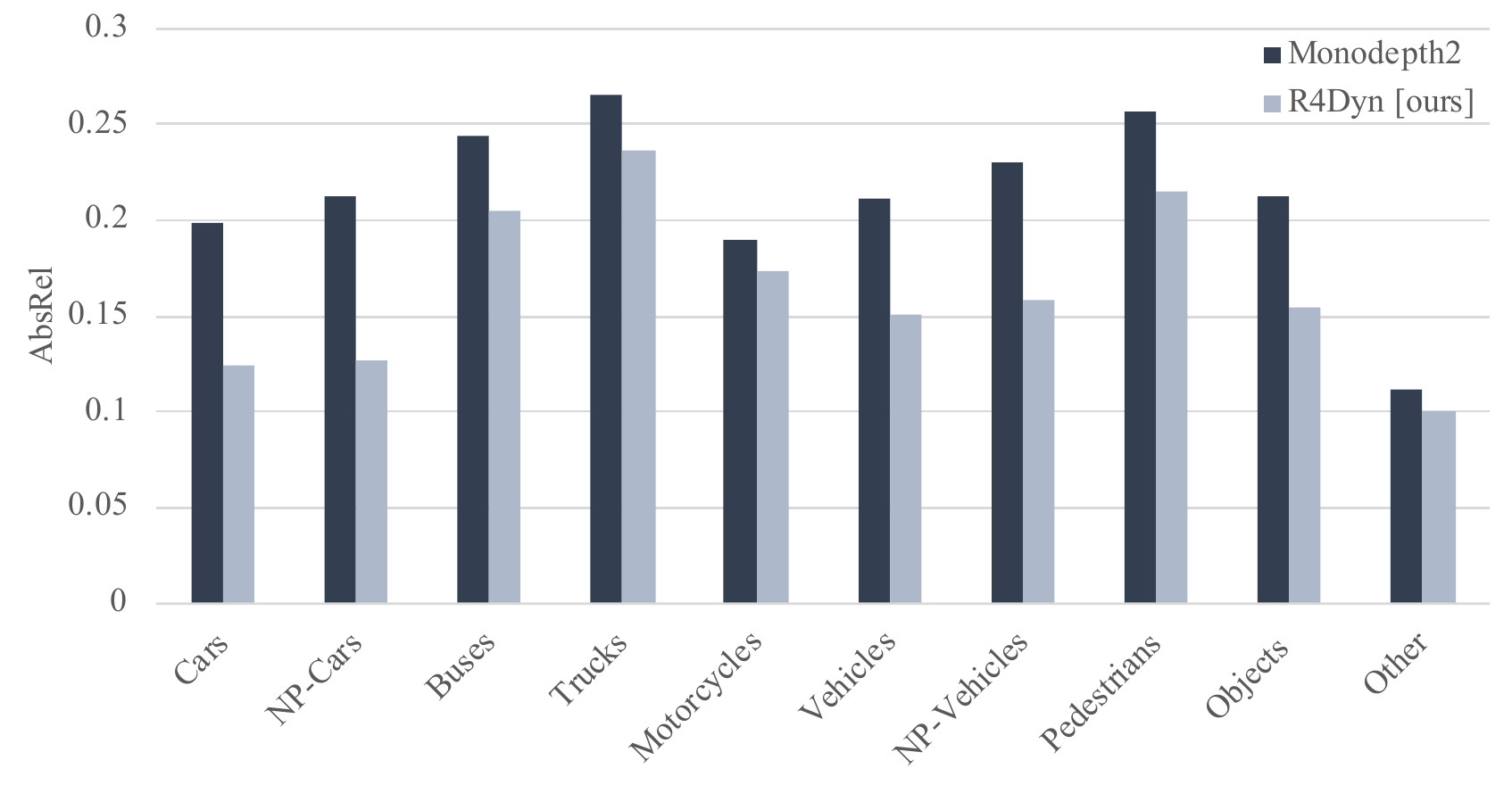}
   \caption{The histogram shows a class-wise evaluation of our R4Dyn and Monodepth2~\cite{2019-godard} on the nuScenes \textit{day-clear} validation set~\cite{2020-caesar}. \textit{Other} includes all remaining classes, that are not traffic participants, and \textit{NP} indicates \textit{Non-Parked} \textit{Cars} or \textit{Vehicles}. Lower absolute relative (AbsRel) error is better.}
   \label{fig:class_eval_histogram}
\end{figure}

\subsection{Additional Results} \label{subsec:additional_quantitative_results}

\subsubsection{Qualitative Ablation Studies}

In Figures~\ref{fig:qual_loss_ablation} and~\ref{fig:qual_input_ablation} we show highlights of the qualitative results of our ablation studies on the weak radar supervision and additional radar input, respectively. These complement the quantitative results reported in Tables~\ref{table:ablation_loss_function} and~\ref{table:input_ablation} with the sample being the same as the first one in Figure~\ref{fig:method_comparison} for consistency. In the Figures, it can be seen how our image-only baseline (L1 and I1) performs similarly to Monodepth2~\cite{2019-godard} on which it is based. Then, in Figure~\ref{fig:qual_loss_ablation}, adding a weak supervision by raw radar points (L2) deteriorated the estimations due to the considerable amount of noise caused by multi-path effects~\cite{2021-long}, resulting in an overestimated set of pixels around the horizon line, which is the image location where most radar detections lie on (as can be seen in Figure~\ref{fig:qual_loss_ablation}).
As reported in Table~\ref{table:ablation_loss_function}, filtering the radar points (L3) led to a significant improvement in the estimations, which then improved further with our R4Dyn-L (L6), e.g. for the leading vehicle in the image.

For the input ablation (Figure~\ref{fig:qual_input_ablation}), a single frame of radar input measurements (I2) did not improve the predictions, while the weak radar supervision and accumulating multiple radar measurements (I4) eliminate the severe underestimation, but delivers artifacts on the shape of vehicles (e.g. oncoming van), which are improved by R4Dyn-LI (I6).

\subsubsection{Class-wise Comparison with Related Works} \label{subsubsec:class_wise_evaluation}
Additionally to the absolute relative error (AbsRel) on object classes \textit{Cars}, \textit{Vehicles}, \textit{Non-parked Vehicles} and \textit{Pedestrians} reported in Table~\ref{table:eval_all}, we provide extensive class-wise results in Tables~\ref{tab:class_eval_vehicles} and~\ref{tab:class_eval_remaining}. Moreover, we plot a comparison between Monodepth2~\cite{2019-godard} and our R4Dyn in Figure~\ref{fig:class_eval_histogram} across the various classes. From the Figure, it can be seen that our approach outperformed Monodepth2 on every object class by a significant margin, which was especially large for the most dynamic ones, namely \textit{Cars}, \textit{Vehicles} and their \textit{Non-Parked} (\textit{NP}) variants. Considering Tables~\ref{tab:class_eval_vehicles} and~\ref{tab:class_eval_remaining}, the LiDAR-supervised work by Lin et al.~\cite{2020-lin}, which uses radar as input, was able to deliver superior estimates on non-object classes, denoted as \textit{Other} (e.g. \textit{Driveable Surface} and \textit{Vegetation}).
Nevertheless, our R4Dyn could obtain significantly lower errors and better scores on all object classes, except for \textit{Trucks}, probably due to trucks being often static in urban areas (e.g. for loading and for deliveries), such as in the nuScenes dataset~\cite{2020-caesar}. The results re-emphasize and confirm the effectiveness and the robustness of our approach across various safety critical dynamic objects, as well as the benefit of using radar sensors for depth estimation, especially as weak supervision.

\subsubsection{Class-wise Comparison on Adverse Conditions}
\label{subsubsec:class_wise_evaluation}
In Table~\ref{table:generalizability_vehicles} we provide class-wise results of ours and related methods under adverse weather settings. As for Table~\ref{table:generalizability}, which reports general errors in the same weather conditions, all approaches were trained on \textit{day-clear} scenes (indicated as \textit{clear} in the Table), therefore the values represent the ability of each method to generalize to rather different inputs. In particular, in Table~\ref{table:generalizability_vehicles} we provide AbsRel errors on safety critical traffic participants: \textit{Cars}, \textit{Vehicles} and \textit{Pedestrians}.
For these classes, our R4Dyn-LI was able to outperform related methods by a significant margin, under most settings. Again, we attribute this to the benefit of radar and the effectiveness of our techniques to incorporate it.

\begin{table}[h]
\begin{center}
    \begin{tabular}{l|ll|c}
    \toprule 
    Method & Supervision & Input & AbsRel  \\
    \midrule
    Monodepth2~\cite{2019-godard} & nuScenes: M* & Im & 0.1162\\
    R4Dyn~[ours] & nuScenes: Mvr & Im$\Psi$ & \textbf{0.1028} \\
    \bottomrule
    \end{tabular}
\end{center}
\caption{Transfer comparison to the KITTI validation set of models trained on nuScenes. $\Psi$ indicates LiDAR sub-sampled like a radar.}
\label{tab:kitti}
\end{table}

\subsubsection{Transfer Comparison: nuScenes $\rightarrow$ KITTI}
\label{subsubsec:transfer_evaluation}
As written in Section~\ref{subsec:exp_setup}, we used the nuScenes dataset~\cite{2020-caesar} throughout our experiments, as it is to the best of our knowledge the only public dataset with data from a camera, an automotive radar, and a LiDAR to evaluate. On the other hand, the popular KITTI dataset~\cite{2013-geiger} does not provide radar signals and has only limited dynamic scenes, which are the focus of our work. Nevertheless, radar sparsity and missing elevation can be simulated by sub-sampling a LiDAR signal, albeit leaving behind the highly complex radar noise modelling~\cite{2021-long}, as done also by Lin et al.~\cite{2020-lin}.
Towards this end, we transferred our R4Dyn and Monodepth2~\cite{2019-godard} models to the validation set of KITTI, without any fine-tuning, after training them on nuScenes~\cite{2020-caesar}. We achieved this by cropping the KITTI images to the nuScenes format, and downsampled the KITTI LiDAR to mimic the nuScenes radar: we kept only points within the radar field of view and sub-sampled them to reach the same low density (Section~\ref{subsec:method_radar_loss_expansion}).

\begin{figure}[h]
\begin{center}
\includegraphics[width=0.47\textwidth]{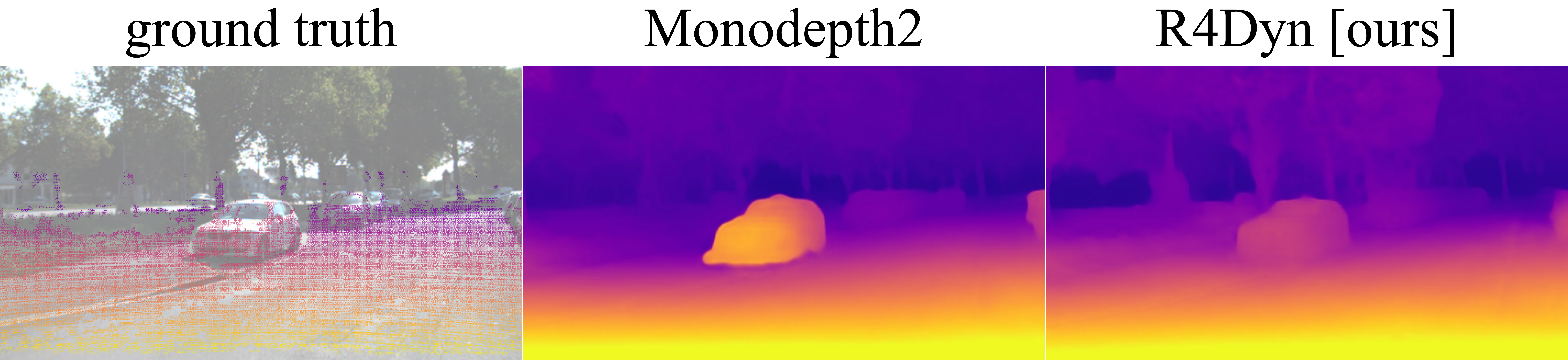}
\end{center}
\caption{Qualitative example from a transfer from nuScenes to KITTI, without any fine-tuning of our R4Dyn, compared to Monodepth2~\cite{2019-godard}.}
\label{fig:kitti_qualitative}
\end{figure}

We provide the results of these experiments in Table~\ref{tab:kitti}, showing the generalization capability on significantly different data.
Our R4Dyn outperforms Monodepth2 in this challenging transfer task. Qualitative results in Figure~\ref{fig:kitti_qualitative} confirm this and present similar outcomes to those seen in Figure~\ref{fig:method_comparison}, with Monodepth2 underestimating the depth of the oncoming vehicle. Despite the large domain gap and limited dynamic scenes (i.e.~our focus), our 11.5\% improvement on KITTI is substantial, again showing the effectiveness of our techniques to use radar in input and as weak supervision.

{\small
\bibliographystyle{ieee_fullname}
\bibliography{egbib}
}



\end{document}